\documentclass[final,3p,times]{elsarticle}

\usepackage[colorlinks=true]{hyperref}

\usepackage{amssymb}

\usepackage{lipsum}
\usepackage[labelsep=period,singlelinecheck=off,textformat=period]{caption}
\usepackage{subcaption}
\usepackage{fancyhdr} 
\usepackage{epsfig}
\usepackage{timesmt}
\usepackage{mathtools,nccmath}
\usepackage{mathrsfs}
\usepackage{wrapfig}
\usepackage[noabbrev]{cleveref}
\creflabelformat{equation}{#2\textup{#1}#3}

\usepackage{pdflscape}
\usepackage{afterpage}
\usepackage{float}
\usepackage{url}
\usepackage{enumitem}
\usepackage{accents}
\usepackage{cuted}
\usepackage{dirtytalk}
\usepackage{multirow}
\usepackage{multicol}
\usepackage{booktabs}
\usepackage{array}
\usepackage{makecell}
\usepackage{colortbl}
\usepackage{soul}
\newcolumntype{n}{>{\columncolor{blue!5}}c}

\usepackage{pbox}
\usepackage[ruled,commentsnumbered,linesnumbered,boxed]{algorithm2e}
\SetArgSty{textnormal}
\SetKwComment{Comment}{/* }{ */}

\usepackage[flushleft]{threeparttable}
\usepackage{algpseudocode}
\usepackage{setspace}
\usepackage{courier}
\usepackage{psfrag,pstool}

\usepackage[normalem]{ulem}

\newcommand{\comment}[1]{}

\Crefname{equation}{Eq.}{Eqs.}
\Crefname{figure}{Fig.}{Figs.}
\Crefname{tabular}{Tab.}{Tabs.}
\crefname{algocf}{alg.}{algs.}
\Crefname{algocf}{Algorithm}{Algorithms}
\Crefname{fct}{Fact}{Facts }

\usepackage{amsmath}
\usepackage{amssymb}
\usepackage{amsthm}
\usepackage{mathrsfs}

\usepackage{bm}
\usepackage{mathtools}
\usepackage{nicefrac}

\theoremstyle{plain}

\theoremstyle{definition}

\theoremstyle{remark}
\newtheorem{rem}{Remark}

\newcommand{\transpose}{^{\mathrm{T}}}

\newcommand{\probb}[2]{\mathrm{p}_{#1}\!\left({#2}\right)}
\newcommand{\prob}[1]{\mathrm{p}_{#1}}

\newcommand{\latin}[1]{\textit{#1}}
\renewcommand{\Re}{\mathbb{R}}

\newcommand{\am}{{\mathbf a}}

\newcommand{\Bm}{{\mathbf B}}
\newcommand{\bmm}{{\mathbf b}}

\newcommand{\Exp}{\mathbb{E}}

\newcommand{\Fii}{\mathit{F}}

\newcommand{\Ind}{\mathbb{I}}

\newcommand{\M}{{ \mu}}

\newcommand{\Ns}{n_\mathrm{s}}

\newcommand{\sm}{\mathbf{s}}

\newcommand{\uu}{{\mathbf u }}

\newcommand{\x}{{\mathbf x }}

\newcommand{\X}{{\mathbf X}}

\newcommand{\y}{{\mathbf y}}
\newcommand{\Y}{{\mathbf Y}}

\newcommand{\z}{{\mathbf z}}
\newcommand{\Z}{{\mathbf Z}}

\newcommand{\Sigm}{\boldsymbol{\Sigma}}

\newcommand{\thetaa}{\bm{\theta}}

\newcommand{\Nx}{{n_{\mathcal{X}}}}
\newcommand{\Ny}{{n_{\mathcal{Y}}}}

\newcommand{\data}{\hat{\y}}
\newcommand{\noise}{\bm{\eta}}

\newcommand{\ie}{i.e., }

\newcommand{\supth}[1]{\ensuremath{{#1}^{\text{th}}}}

\newcommand{\pwr}[2]{{#1}$\times$10\textsuperscript{#2}}

\usepackage{tikz}

\usetikzlibrary{shapes.misc}
\usetikzlibrary{arrows,arrows.meta,calc,backgrounds}

\tikzstyle{block} = [draw,rectangle,thick,minimum height=2em,minimum width=2em]
\tikzstyle{sum} = [draw,circle,inner sep=0mm,minimum size=2mm]
\tikzstyle{connector} = [->,thick]
\tikzstyle{line} = [thick]
\tikzstyle{branch} = [circle,inner sep=0pt,minimum size=1mm,fill=black,draw=black]
\tikzstyle{guide} = []
\tikzset{>=latex}

\makeatletter
\newcommand{\hathat}[1]{%
	\begingroup%
	\let\macc@kerna\z@%
	\let\macc@kernb\z@%
	\let\macc@nucleus\@empty%
	\hat{\raisebox{.35ex}{\vphantom{\ensuremath{#1}}}\smash{\hat{#1}}}%
	\endgroup%
}
\makeatother

\creflabelformat{equation}{#2(#1)#3}
\crefrangelabelformat{equation}{#3(#1)#4 to #5(#2)#6}

\labelcrefformat{subequation}{#2(#1)#3}
\labelcrefrangeformat{subequation}{#3(#1)#4 to #5(#2)#6}

\renewcommand{\mathbf}[1]{\bm{#1}}

\usepackage{lineno}

\journal{Computer Methods in Applied Mechanics and Engineering}

\begin{document}
	
	\begin{frontmatter}
		
		\title{Conditional score-based diffusion models for solving inverse elasticity problems}
		
		\author[label1]{Agnimitra Dasgupta}
		\author[label1]{Harisankar Ramaswamy}
		\author[label1]{Javier Murgoitio-Esandi}
		\author[label2,label3,label4]{Ken Foo}
		\author[label5]{Runze Li}
		\author[label6]{Qifa Zhou}
		\author[label2,label3,label4,label7]{Brendan F Kennedy}
		\author[label1]{Assad A Oberai\corref{cor1}}
		
		\cortext[cor1]{Corresponding author}
		\ead{aoberai@usc.edu}

		\affiliation[label1]{organization={{Department of Aerospace \& Mechanical Engineering, University of Southern California}},
			city={Los Angeles},
			postcode={90089}, 
			state={California},
			country={USA}}
		
		\affiliation[label2]{organization={{BRITElab, Harry Perkins Institute of Medical Research, QEII Medical Centre}},
			city={Nedlands},
			postcode={6009}, 
			state={Western Australia},
			country={Australia}}
		
		\affiliation[label3]{organization={{Centre for Medical Research, The University of Western Australia}},
			city={Perth},
			postcode={6009}, 
			state={Western Australia},
			country={Australia}}

		\affiliation[label4]{organization={{Department of Electrical, Electronic \& Computer Engineering, School of Engineering, The
					University of Western Australia}},
			city={Perth},
			postcode={6009}, 
			state={Western Australia},
			country={Australia}}
		
		\affiliation[label7]{organization={{Institute of Physics, Faculty of Physics, Astronomy and Informatics, Nicolaus Copernicus University in Toruń}},
			city={Grudziadzka 5},
			postcode={87-100}, 
			state={Toruń},
			country={Poland}}
		
		\affiliation[label5]{organization={{Department of Mechanical Engineering, Massachusetts Institute of Technology}},
			city={Cambridge},
			postcode={02139}, 
			state={Massachusetts},
			country={USA}}
		
		\affiliation[label6]{organization={{Alfred E. Mann Department of Biomedical Engineering, University of Southern California}},
			city={Los Angeles},
			postcode={90089}, 
			state={California},
			country={USA}}	
		
		\begin{abstract}
			We propose a framework to perform Bayesian inference using conditional score-based diffusion models to solve a class of inverse problems in mechanics involving the inference of a specimen's spatially varying material properties from noisy measurements of its mechanical response to loading. Conditional score-based diffusion models are generative models that learn to approximate the score function of a conditional distribution using samples from the joint distribution. More specifically, the score functions corresponding to multiple realizations of the measurement are approximated using a single neural network, the so-called score network, which is subsequently used to sample the posterior distribution using an appropriate Markov chain Monte Carlo scheme based on Langevin dynamics. Training the score network only requires simulating the forward model. Hence, the proposed approach can accommodate black-box forward models and complex measurement noise. Moreover, once the score network has been trained, it can be re-used to solve the inverse problem for different realizations of the measurements. We demonstrate the efficacy of the proposed approach on a suite of high-dimensional inverse problems in mechanics that involve inferring heterogeneous material properties from noisy measurements. Some examples we consider involve synthetic data, while others include data collected from actual elastography experiments. Further, our applications demonstrate that the proposed approach can handle different measurement modalities, complex patterns in the inferred quantities, non-Gaussian and non-additive noise models, and nonlinear black-box forward models. The results show that the proposed framework can solve large-scale physics-based inverse problems efficiently.
		\end{abstract}
		
		
		
		
	\end{frontmatter}
	
	\section{Introduction}\label{sec:introduction}
	
	Generative artificial intelligence (AI) is at the forefront of the latest wave of advances in deep learning technology. Generative AI is the backbone for much of the technology that has pervaded our society, such as large language models and AI-driven image and video generators. Generative models seek to understand how data is generated and replicate the data generation process to synthesize new data as desired. The essence of generative models is as follows: given data sampled from an underlying probability distribution, inaccessible at the outset, generative models learn to sample new realizations from the same probability distribution~\cite{pml2Book}. Mathematically, a generative model is a means to approximate the underlying probability distribution, say $\prob{\X}$, using realizations of $\X$ that are available at hand. Several types of generative models have been developed, including generative adversarial networks (GANs), variational autoencoders (VAEs), autoregressive models (ARMs), normalizing flows (NFs), energy-based models (EBMs), and diffusion models. These models vary in the way they approximate the underlying probability distribution and whether they construct an explicit or implicit approximation~\cite{pml2Book}. Among them, diffusion models are emerging as the leading choice and have achieved state-of-the-art performance in image synthesis~\cite{dhariwal2021diffusion}. Indeed, diffusion models are used by the latest generation of AI-based text, image, and video generators~\cite{gpt4tech2023,ramesh2022hierarchical,esser2024scaling}.

	Sometimes, the generation process must be conditioned on inputs or covariates, say $\Y$. For instance, when generating images based on text prompts~\cite{ramesh2022hierarchical}. Mathematically framed, a conditional generative model must learn the conditional distribution $\prob{\X\vert\Y}$ using pairwise realizations from the underlying joint distribution of $\X$ and $\Y$. In particular, conditional diffusion models are increasingly being used across several fields in science and engineering; see recent reviews~\cite{yang2023diffusion,li2023diffusion,croitoru2023diffusion,kazerouni2022diffusion,guo2024diffusion}. Some notable applications in computational sciences and engineering include material microstructure generation based on various microstructural descriptors~\cite{dureth2023conditional} and desired nonlinear behavior~\cite{vlassis2023denoising}, metamaterial synthesis to obtain specified stress-strain response~\cite{bastek2023inverse}, topology optimization under constraints~\cite{maze2023diffusion}, reconstruction of turbulent flow fields~\cite{li2023multi}, and synthesizing high-fidelity simulations of turbulent flow from low-fidelity data~\cite{shu2023physics}.

	Clearly, conditional generative modeling resembles the process of solving an inverse problem within the Bayesian paradigm where $\X$ and $\Y$ now represent the quantity that must be inferred and the measurements available, respectively~\cite{dimakis2022deep}. Thus, motivated by the recent wide-scale adoption of diffusion models, we proposed to use conditional diffusion models to solve a class of inverse problems in mechanics that involve inferring the spatially varying material constitutive parameters from full-field measurements of the mechanical response, such as displacements or strains. Inverse problems of this type have many applications in material characterization~\cite{avril2008overview}, structural health monitoring and damage detection~\cite{yang2017full,dong2020structural}, and medical diagnosis~\cite{sarvazyan1998mechanical,patel2019circumventing}. We are particularly motivated by the applications of this type of inverse problem in medical diagnosis, which arises when using elastography-based imaging techniques that assess the state of tissue using its mechanical properties, mainly to differentiate between healthy and diseased tissue~\cite{sarvazyan1998mechanical,barbone2007elastic}. 
	
	Solving inverse problems of the aforementioned type can be challenging. First, the inferred quantity is a field that must be discretized to make the inverse problem tractable, leading to a high-dimensional inverse problem. Second, the forward model is often complex. Mechanistic models that simulate the deformations of the specimen in consideration may be a high-fidelity black-box model. Moreover, the measurement noise in these applications may be complex--- non-Gaussian and non-additive. These factors combine to make Bayesian inference difficult since standard inference tools, such as Markov chain Monte Carlo (MCMC) methods and variational inference, may not perform well in the face of these challenges. However, the recent success of conditional diffusion models in solving high-dimensional inverse problems offers promise~\cite{song2021solving,song2022pseudoinverse,chung2022improving,chung2022diffusion,chung2022come}. 
	
	We propose to use conditional score-based diffusion models (cSDMs) --- a particular type of diffusion model that leverages deep neural networks to approximate the \emph{score function} of the target posterior distribution~\cite{hyvarinen2005estimation,vincent2011connection,song2019generative}. In the context of inverse problems, the score function is the logarithm of the posterior density. The neural network approximating the score function is called the \textit{score network}~\cite{song2019generative}. Approximating the score function allows greater flexibility in choosing the architecture for the score network in the absence of additional constraints necessary to model a valid probability density. For instance, the score network need not be invertible~\cite{song2019generative}, unlike normalizing flows~\cite{kobyzev2020normalizing}. Moreover, a single neural network can approximate the posterior distribution for different realizations of the covariate $\Y$, which only requires adding a channel to the input of the neural network~\cite{saharia2022image,song2020score,batzolis2021conditional}. This also allows for re-using the trained score network for different realizations of $\Y$ during testing. Significantly, we show that the score network can be trained using a regression-type objective, see \Cref{eq:eDSM-fianl}, which is much more stable than adversarial losses used to train GANs. Moreover, training only requires samples from the joint distribution between $\X$ and $\Y$. The training dataset can be easily created by simulating the forward model for different realizations of $\X$ sampled from a suitable prior or collected a priori through experiments. Thus, the forward model can be arbitrarily complex, black-box, and incompatible with automatic differentiation. This lends great utility to the proposed approach since most forward models for the inverse problems of practical interest are likely to have one or more of the aforementioned properties. Moreover, the training dataset is entirely synthetic and does not require access to experimental data, which can be scarce in some applications. After training the score network, we can sample the posterior distribution for any realization of $\Y$, perhaps the result of an experiment, using Langevin dynamics-based MCMC~\cite{roberts1998optimal,song2019generative,song2020improved}.

	We note that recent work by \citet{jacobsen2023cocogen} and \citet{huang2024diffusionpde} uses diffusion models for full-field inversion from sparse measurements. \citet{jacobsen2023cocogen} primarily focuses on enforcing physical consistency to the realizations generated using the conditional diffusion model, which is not the focus of the current work. Similarly, \citet{huang2024diffusionpde} train score-based diffusion models for unconditional generation on the joint space between the full-field coefficients and the corresponding solution. Measurement mismatch error and physics-informed residue terms help guide the sampling process to simultaneously recover missing observations and perform full-field inversion. The realizations obtained using cSDMs can be enhanced by enforcing physical consistency using ideas developed in \cite{jacobsen2023cocogen,huang2024diffusionpde}. In contrast, \citet{bastek2024physics} propose to enforce physical consistency during training of the score network, which can also be used to augment the proposed approach. However, enforcing physical consistency by discretizing the underlying forward model may not be straightforward in many applications involving nonlinear phenomena and complex geometries. Therefore, we do not pursue informing the score network or the sampling process with physics knowledge in this work. Indeed, the proposed approach is agnostic to the physics of the forward process, \ie we assume that the forward model is black-box. 
	
	We also note that deep generative models have been used to solve inverse problems~\cite{dimakis2022deep,ongie2020deep}. One line of work uses deep generative models as data informative priors for solving inverse problems~\cite{patel2021solution,bohra2022bayesian,whang2021composing}. Similar to this work, conditional generative models constructed using GANs~\cite{adler2018deep,ray2022efficacy,ray2023solution} and NFs~\cite{padmanabha2021solving} have been used to sample the posterior in inverse problems. NFs have also been used for variational Bayesian inference~\cite{sun2021deep,dasgupta2021uncertainty}. More recently, modular Bayesian frameworks have also been developed that use coupled generative models to perform Bayesian inference~\cite{feng2023dimension,dasgupta2024dimension}. Compared to diffusion models, other generative models suffer from various limitations, such as mode collapse in GANs, poor synthesis quality in VAEs, and the need for specialized architecture with large memory footprints in NFs~\cite{saharia2022image}. Thus, there is now a growing body of work that utilizes diffusion models to solve inverse problems~\cite{song2022pseudoinverse,graikos2022diffusion,song2021solving,chung2022improving,chung2022diffusion,mardani2023variational}. These works use the gradient of the likelihood function to influence the sample generation process. However, computing the gradient of the likelihood function usually requires taking gradients of the black-box nonlinear forward model, which can be computationally expensive. Thus, efforts have been made to approximate the gradient of the likelihood function. In contrast, we use the score network to directly approximate the score function, similar to \cite{saharia2022image}. Whereas \citet{saharia2022image} considers the problem of image super-resolution, we are motivated by inverse problems arising in solid mechanics. 
	
	The remainder of this paper is organized as follows. In \Cref{sec:background}, we set up the inverse problem we intend to solve and provide the necessary background on score-based diffusion models. We introduce our proposed approach to solve inverse problems using cSDMs in \Cref{sec:proposedapproach} where \Cref{subsec:training} develops the training objective, \Cref{subsec:sampling} outlines the sampling scheme, \Cref{subsec:hyperparameters} provides guidelines to choose various hyper-parameters associated with the model and sampling, \Cref{subsec:design-score} provides design details pertaining to the neural network used to approximate the score function, and \Cref{subsec:workflow} provides an overview of the various steps in the proposed approach. In \Cref{sec:results}, we use the proposed approach to solve various inverse problems that concern the inference of a spatially varying constitutive parameter from noisy measurements of sample deformation. The applications include high-dimensional inverse problems with synthetic (\Cref{subsec:synthetic}) and experimental data (\Cref{subsec:experimental}). \Cref{sec:conclusion} ends with conclusions and future directions.

	\section{Background} \label{sec:background}
	
	\subsection{Problem setup}
	A typical inverse problem in mechanics involves inferring the spatially varying material constitutive parameter(s) of a specimen from measurements, likely noisy and possibly sparse, of its mechanical response under some controlled loading. The material constitutive parameter(s) that must be inferred is a continuous field quantity and discretizing it helps make the inverse problem tractable. We denote using $\X$ the $\Nx$-dimensional vector of material constitutive parameter(s) that has to be inferred. The \supth{i} component of $\X$ denotes the material constitutive parameter at some unique physical coordinate on the specimen. Similarly, we denote using $\Y$ the $\Ny$-dimensional prediction vector of the specimen's response. Associated with the inverse problem is a mechanistic model $\Fii$, which we can evaluate to obtain the prediction for a given input $\X$. $\Fii$ is likely to be a computational model which we can query to obtain the response of a specimen for a given input spatial distribution of the material constitutive parameter $\x$, and initial and boundary conditions. This computational model can be a high-fidelity black-box model, making it incompatible with automatic differentiation and, therefore,  impossible to interface with deep learning libraries. It can also be probabilistic, thereby accounting for the effect of the stochastic measurement moise and model error. Finally, the goal of the inverse problem is to infer $\X$ from a measurement $\hat{\y}$, a realization of $\Y$, corrupted with noise. 
	
	\subsection{Bayesian inference}
	
	Bayesian inference is a popular framework for solving inverse problems in a probabilistic setting. The Bayesian treatment of an inverse problem begins with the designation of $\X$ and $\Y$ as random vectors. Herein, $\prob{\X}$ denotes the density of the probability distribution of $\X$; the latter is popularly know as the \emph{prior distribution}. Now, let $\prob{\Y \vert \X}$ denote the density corresponding to the probability distribution of $\Y$ conditioned on $\X$, also known as the \emph{likelihood} distribution, induced by the forward model
	\begin{equation}\label{eq:forward-model}
		\y = \mathcal{F}(\x; \noise),
	\end{equation}
	where $\x$ and $\y$ denote realizations of $\X$ and $\Y$, respectively, and $\noise$ is a random vector representing measurement noise and modeling errors. Drawing a realization of $\Y$ given a realization of $\X$ using \Cref{eq:forward-model} involves evaluating the computational model $\Fii$ and sampling $\noise$ from an appropriate distribution. Herein, we assume that we have access to the likelihood, which implies that we can sample $\noise$. Now, given the  measurement $\hat{\y}$, which is a realization of the random vector $\Y$, the goal of Bayesian inference is to obtain the conditional probability distribution, also known as the \emph{posterior} distribution, with density $\prob{\X\vert\Y}\!\left(\cdot \vert \Y = \hat{\y}\right)$ as follows:
	\begin{equation}\label{eq:Bayes}
		\prob{\X\vert\Y}\!\left(\X = \x \vert \Y = \hat{\y}\right) \propto \prob{\Y\vert\X}\!\left(\Y = \hat{\y} \vert \X = \x \right) \probb{\X}{\X = \x}.
	\end{equation}
	In the remainder of this manuscript, we drop the random variables in the argument of the respective densities to simplify notation. For instance, $\probb{\X}{\x}$ will mean $\probb{\X}{\X = \x}$. 
	
	As simple as the Bayes' rule \Cref{eq:Bayes} is, its application can be very challenging. We cannot evaluate the posterior analytically unless the prior and the likelihood distributions form a conjugate pair. A conjugate prior and likelihood pair is unlikely to arise in applications relevant to the current work since the noise may be non-additive and the computational model  $\Fii$ may be non-linear. Hence, we must approximate the posterior distribution. One way is to draw samples from the posterior distribution, which we can use to compute posterior statistics and posterior predictive quantities. However, sampling high-dimensional posteriors is far from straightforward. 
	
	\subsection{Langevin dynamics and the score function}
	
	MCMC algorithms have been the staple of Bayesian machinery. MCMC methods are useful for sampling unnormalized probability models such as the posterior distribution in inverse problems. Early MCMC methods based on the Metropolis-Hastings (MH) algorithm \cite{metropolis1953equation, hastings1970monte} use a proposal density to advance a Markov chain with an invariant distribution similar to the target posterior. However, in high-dimensional settings, simple MH-type MCMC methods are known to be inefficient as randomly chosen new states fail to explore high-likelihood regions quickly.
	
	Langevin dynamics is an improvement over MH-MCMC, and uses a proposal density obtained from discretizing Langevin diffusion equation with a drift term informed by the gradient of the target posterior distribution~\cite{roberts1998optimal}. The resulting stochastic differential equation is defined as:
	\begin{equation}\label{eq:langevin}
		\mathrm{d} \X_t = \frac{\nabla_{\x} \log \probb{\X \vert \Y}{\x \vert \hat{\y}} }{2} \mathrm{d}t + \mathrm{d}\Bm_t,
	\end{equation}
	where $\Bm$ represent as $\Nx$-dimensional Brownian motion. The first term on the right hand side of \Cref{eq:langevin} is popularly known as the \emph{score} function and it has two noteworthy properties. First, the score function is vector valued, \ie $\nabla_{\x} \log \probb{\X \vert \Y}{\x \vert \hat{\y}} : \Re^{\Nx} \to \Re^{\Nx}$.  Second, the score function does not depend on the normalization constant of the posterior distribution, \ie 
	\begin{equation}
		\probb{\X \vert \Y}{\x \vert \hat{\y}} \propto \mathrm{q}_{\X \vert \Y}\!\left(\x \vert \hat{\y}\right) \implies \nabla_{\tilde{\x}} \log \probb{\X \vert \Y}{\x \vert \hat{\y}} = \nabla_{\x} \log \mathrm{q}_{\X \vert \Y}\!\left(\x \vert \hat{\y}\right) ,
	\end{equation}
	where the density $\mathrm{q}_{\X \vert \Y}\!\left(\x \vert \hat{\y}\right) $ is unnormalized, \ie $\int \mathrm{q}_{\X \vert \Y}\!\left(\x \vert \hat{\y}\right)  \; \mathrm{d}\x \neq 1$. Further, a first order Euler-Maruyama discretization of \Cref{eq:langevin} yields the following iterative proposal mechanism: 
	\begin{equation}\label{eq:langevin-discrete}
		\x_{t+1} = \x_t + \frac{\epsilon}{2}  \nabla_{\x} \log \probb{\X \vert \Y}{\x_{t} \vert \hat{\y}}  + \sqrt{\epsilon} \z_t,
	\end{equation}
	where $\epsilon$ is the integration step size and $\z_t$ are independent and identical realizations of standard normal variable $\Z$. 
	\Cref{eq:langevin-discrete} converges to the invariant distribution $\prob{\X \vert \Y}$ as time $t \to \infty$. As evident from \Cref{eq:langevin,eq:langevin-discrete}, the score function $\nabla_{\x} \log \prob{\X \vert \Y}$ is crucial for sampling the posterior distribution using Langevin dynamics. 
	
	\section{Bayesian inference using conditional score-based diffusion models} \label{sec:proposedapproach}
	
	Bayesian inference methods try to approximate the posterior distribution of $\X$ for a given measurement $\hat{\y}$ using samples drawn from it. \Cref{eq:langevin,eq:langevin-discrete} are useful for devising an MCMC method for sampling the posterior distribution. However, the `\emph{true}' score function remains inaccessible in practical applications. Using cSDMs, we approximate the score function using a neural network $\sm(\x, \y; \thetaa)$, popularly known as the \emph{score network}, where $\thetaa$ represents the parameters of the neural network. In this section, we develop the training objective for the score network, introduce the MCMC-based sampler called Annealed Langevin dynamics~\cite{song2019generative} that is useful for sampling target posteriors using a trained score network, provide guidelines regarding hyper-parameter selection and design choices for the score network, and finally provide an overview of the entire workflow. 
	
	\subsection{Training the score network using denoising score matching}\label{subsec:training}
	
	Consider a realization $\y$ of $\Y$ associated with which is the target posterior distribution $\probb{\X\vert\Y}{\x \vert \y}$ with the score function $\nabla_{\x} \log \probb{\X\vert\Y}{\x \vert \y}$. We want the score network $\sm(\x, \y; \thetaa)$ to be a good approximation to the score function over the entire support of $\X$. So, a potential candidate objective function to train the score network is the score matching objective, 
	\begin{equation}\label{eq:esm}
		\ell_{\mathrm{SM}}(\y; \thetaa) = \frac{1}{2}\Exp_{\X \sim \prob{\X\vert\Y}} \Big[ \lVert  \sm(\x, \y; \thetaa) - \nabla_{\x} \log \probb{\X\vert\Y}{\x\vert\y}\rVert_2^2 \Big].
	\end{equation}
	One can show that, under some mild regularity conditions, the score matching objective in \Cref{eq:esm} is equivalent, up to  a constant, to the following 
	\begin{equation}\label{eq:esm-trace}
		\ell_{\mathrm{SM}}(\y; \thetaa) =  \Exp_{\X \sim \prob{\X\vert\Y}} \Big[ \frac{1}{2} \lVert  \sm(\x, \y; \thetaa) \rVert_2^2 + \mathrm{Tr}(\nabla_{\x} \sm(\x, \y; \thetaa))\Big],
	\end{equation}
	where $\mathrm{Tr}(\cdot)$ is the trace operator~\cite{hyvarinen2005estimation,song2019generative}. However, \Cref{eq:esm-trace} is not scalable to high-dimensional problems since it involves computing the gradient of the output of the score network with respect to its input, the computational cost of which will scale linearly with the dimension $\Nx$ of the inverse problem. To circumvent this issue, \citet{hyvarinen2005estimation} proposed a \emph{denoising score matching} objective that altogether bypasses computing $\nabla_{\x} \sm(\x, \y; \thetaa)$. 
	
	Denoising score matching avoids the computation of $\nabla_{\x} \sm(\x, \y; \thetaa)$ by first perturbing the random variable $\X$, conditioned on $\Y$, to obtain $\tilde{\X}$ using a perturbation kernel $\prob{\tilde{\X}\vert\X}$. Note that this makes $\tilde{\X}$ and $\Y$ conditionally independent given $\X$. Now, the denoising score matching objective, for a given realization $\y$ of $\Y$, is defined as
	\begin{equation}\label{eq:dsm1}
		\ell_{\mathrm{DSM}}(\y; \thetaa) =  \frac{1}{2}\Exp_{\tilde{\X} \sim \prob{\tilde{\X}\vert\Y}} \Big[ \lVert  \sm(\tilde{\x}, \y; \thetaa) - \nabla_{\tilde{\x}} \log  \prob{\tilde{\X}\vert\Y}\!(\tilde{\x}\vert\y)\rVert_2^2 \Big].
	\end{equation}
	Using the fact that
	\begin{equation}\label{eq:convolution}
		\prob{\tilde{\X}\vert\Y}(\tilde{\x} \vert \y) = \int \probb{\tilde{\X}\vert\X}{\tilde{\x}\vert\x} \; \prob{\X\vert\Y}(\x \vert \y) \; \mathrm{d}\x,
	\end{equation}
	one can show that the denoising score matching objective $\ell_{\mathrm{DSM}}$ is equivalent to~\cite{vincent2011connection}:
	\begin{equation}\label{eq:DSM-2}
		\ell_{\mathrm{DSM}}(\y; \thetaa) \equiv  \frac{1}{2} \Exp_{\X \sim \prob{\X\vert\Y}} \Bigg[ \Exp_{\tilde{\X} \sim \prob{\tilde{\X}\vert\X}} \Big[ \lVert  \sm(\tilde{\x}, \y; \thetaa) - \nabla_{\tilde{\x}} \log  \probb{\tilde{\X}\vert\X}{\tilde{\x}\vert\x}\rVert_2^2 \Big] \Bigg],
	\end{equation}
	up to an additive constant that does not depend on $\thetaa$; see \ref{app:derivation} for the derivation. The optimally trained score network with parameters $\thetaa^\ast$, trained using  \Cref{eq:DSM-2}, will satisfy $\sm(\tilde{\x}, \y; \thetaa^\ast) \approx \nabla_{\tilde{\x}} \log  \prob{\tilde{\X}\vert\Y}(\tilde{\x} \vert \y) $ and $\prob{\tilde{\X}\vert\Y}(\tilde{\x} \vert \y)  \approx \prob{\X\vert\Y}(\x\vert \y) $, when the perturbation kernel bandwidth goes to zero. Thus, the training objective \Cref{eq:DSM-2} offers a path to approximating the score function of the target posterior distribution for a specified realization of $\y$ using a score network, provided we choose a very narrow perturbation kernel. 
	
	We note that we wish to use the score network to approximate the score of the posterior distribution over the entire support of $\Y$. Hence, we marginalize $\Y$ out of $\ell_\mathrm{DSM}(\y; \thetaa)$ to obtain the training objective
	\begin{equation}\label{eq:DSM-joint}
		\begin{split}
			\mathcal{L}_{\mathrm{DSM}}(\thetaa) &=   \Exp_{\Y \sim \prob{\Y}} \big[ \ell_{\mathrm{DSM}}(\y; \thetaa) \big] \\
			&\equiv  \frac{1}{2} \Exp_{\Y \sim \prob{\Y}} \Bigg[ \Exp_{\X \sim \prob{\X\vert\Y}} \bigg[ \Exp_{\tilde{\X} \sim \prob{\tilde{\X}\vert\X}} \Big[ \lVert  \sm(\tilde{\x}, \y; \thetaa) - \nabla_{\tilde{\x}} \log  \probb{\tilde{\X}\vert\X}{\tilde{\x}\vert\x}\rVert_2^2 \Big] \bigg] \Bigg] \\
			&= \frac{1}{2} \Exp_{\X, \Y \sim \prob{\X \Y}} \bigg[ \Exp_{\tilde{\X} \sim \prob{\tilde{\X}\vert\X}} \Big[ \lVert  \sm(\tilde{\x}, \y; \thetaa) - \nabla_{\tilde{\x}} \log  \probb{\tilde{\X}\vert\X}{\tilde{\x}\vert\x}\rVert_2^2 \Big] \bigg],
		\end{split}
	\end{equation}
	which is remarkable for one very particular reason. Note, estimating the objective $\mathcal{L}_{\mathrm{DSM}}(\thetaa)$ requires computing two expectations. The outer expectation is with respect to the joint distribution of $\X$ and $\Y$. We can easily sample the joint by first sampling $\X$ from the prior $\prob{\X}$ subsequently $\Y$ from the likelihood $\prob{\Y \vert \X}$, which simply requires the evaluation of the forward model \Cref{eq:forward-model}.  The inner expectation is with respect to the conditional distribution $\prob{\tilde{\X} \vert \X}$, which we can choose to be a tractable distribution that can be easily sampled. Thus, the Monte Carlo (MC) approximation of $\mathcal{L}_{\mathrm{DSM}}(\thetaa)$, which we denote as  $\mathcal{L}_{\mathrm{eDSM}}(\thetaa)$, is given by
	\begin{equation}\label{eq:DSM-empirical}
		\mathcal{L}_{\mathrm{eDSM}}(\thetaa) = \frac{1}{2N_{\mathrm{train}}} \sum_{i=1}^{N_{\mathrm{train}}} \Exp_{\tilde{\X} \sim \prob{\tilde{\X}\vert\X}} \Big[ \lVert  \sm(\tilde{\x}, \y^{(i)}; \thetaa) - \nabla_{\tilde{\x}} \log  \probb{\tilde{\X}\vert\X}{\tilde{\x}\vert\x^{(i)}}\rVert_2^2 \Big],
	\end{equation}
	where $\x^{(i)}$ and $\y^{(i)}$ are the \supth{i} realizations of $\x$ and $\y$, respectively, sampled from the joint $\prob{\X\Y}$.  Moreover, $\y^{(i)} = \mathcal{F}(\x^{(i)}; \noise^{(i)})$, where $\noise^{(i)}$ is the \supth{i} realization of the noise vector. 
	
	There are three practical choices made to aid in the approximation of the score function by the score network. First, the perturbation kernel is assumed to be Gaussian with covariance equal to $\sigma^2\mathbb{I}_{\Nx}$, \ie
	\begin{equation}\label{eq:perturbation_kernel}
		\probb{\tilde{\X}\vert\X}{\tilde{\x}\vert\x} = \mathcal{N}(\tilde{\x}; \x, \sigma^2\mathbb{I}_{\Nx}) \propto \exp \left( -\frac{ \lVert \tilde{\x} - \x \rVert_2^2 }{2\sigma^2} \right) \implies \nabla_{\x} \log \probb{\tilde{\X}\vert\X}{\tilde{\x}\vert\x} =  \frac{\x - \tilde{\x}}{\sigma^2},
	\end{equation}
	where $\sigma$ is referred to as noise level (different from measurement noise) in the literature~\cite{song2019generative,song2020improved} because \Cref{eq:perturbation_kernel} represents adding independent and identically distributed Gaussian noise with variance $\sigma^2$ to a realization $\x$. Thus, the bandwidth of the perturbation kernel $\prob{\tilde{\X}\vert\X}$ scales linearly with $\sigma$. We make the dependence on $\sigma$ explicit by introducing the subscript $\sigma$ to $\tilde{\X}$. For instance, herein, we rewrite $\prob{\tilde{\X}\vert\X}$ as $\prob{\tilde{\X}_{\sigma}\vert\X}$. Second, \citet{song2019generative} suggest that the score function be simultaneously estimated across different noise levels $\{\sigma_i\}_{i=1}^{L}$, where $\sigma_1 > \sigma_2 > \ldots > \sigma_L > 0$. This is akin to `annealing', and, in this case, helps sampling from $\prob{\tilde{\X}\vert\Y}(\tilde{\x} \vert \y)$ by ensuring that less significant modes are smoothed out initially. In fact, sampling will commence from a near uni-modal multivariate Gaussian distribution for a very large value of $\sigma_1$. \citet{song2019generative} also demonstrate that annealing helps in the exploration of low density regions of the joint distribution $\prob{\X\Y}$ when sampling. However, working with $L$ noise levels instead of one also entails a re-parameterization of the score network as $\sm(\tilde{\x}, \y, \sigma; \thetaa)$ as $\sigma$ becomes an additional input to it. However, the input $\sigma$ is used differently from the other inputs; we discuss how the score network uses $\sigma$ in \Cref{subsec:design-score}. Third, the inner expectation in \Cref{eq:DSM-empirical}, with respect to $\Exp_{\tilde{\X} \sim \prob{\tilde{\X}\vert\X}}$, is approximated using a single realization  randomly sampled on the fly during training. 
	
	On assimilating all these practices into \Cref{eq:DSM-empirical}, the new and final training objective becomes
	\begin{equation}
		\mathcal{L}_{\mathrm{eDSM}}(\thetaa) = \frac{1}{2LN_{\mathrm{train}}} \sum_{j=1}^{L} \sum_{i=1}^{N_{\mathrm{train}}}  \lambda(\sigma_j) \Bigg\lVert  \sm(\tilde{\x}_{\sigma_j}^{(i)}, \y^{(i)}, \sigma_j; \thetaa) + \frac{\tilde{\x}_{\sigma_j}^{(i)} - \x^{(i)}}{\sigma_j^2}  \Bigg\rVert_2^2 ,
	\end{equation}
	where $\tilde{\x}_{\sigma_j}^{(i)}$ is the \supth{i} realization of $\tilde{\X}_{\sigma_j}$ sampled from $\prob{\tilde{\X}_{\sigma_j}\vert\X}$ and $\lambda(\sigma_j)$ is a weighting factor chosen equal to $\sigma^2_j$~\cite{song2019generative,song2020improved}, which in turn makes 
	\begin{equation}\label{eq:eDSM-fianl}
		\mathcal{L}_{\mathrm{eDSM}}(\thetaa) = \frac{1}{2LN_{\mathrm{train}}} \sum_{j=1}^{L} \sum_{i=1}^{N_{\mathrm{train}}}  \Bigg\lVert  \sigma_j \sm(\tilde{\x}_{\sigma_j}^{(i)}, \y^{(i)}, \sigma_j; \thetaa) + \frac{\tilde{\x}_{\sigma_j}^{(i)} - \x^{(i)}}{\sigma_j}  \Bigg\rVert_2^2.
	\end{equation}
	A score network of sufficient capacity trained using $\mathcal{L}_{\mathrm{eDSM}}(\thetaa)$, as defined in \Cref{eq:eDSM-fianl}, will approximate the score function across all noise levels. Let $\thetaa^\ast$ denote the optimal parameters of the score network such that
	\begin{equation}
		\thetaa^\ast = \arg \min_{\thetaa} \mathcal{L}_{\mathrm{eDSM}}(\thetaa).
	\end{equation}
	Using the trained score network $\sm(\cdot, \cdot, \cdot; \thetaa^\ast)$, Annealed Langevin dynamics~\cite{song2019generative} can be used to sample from the target posterior distribution. 
	
	\subsection{Sampling using Annealed Langevin Dynamics}\label{subsec:sampling}
	
	From \Cref{eq:convolution}, note, $\prob{\tilde{\X}\vert\Y}$ is the outcome of a convolution between the target posterior $\prob{\X\vert\Y}$ and the zero-mean Gaussian kernel with covariance matrix $\sigma^2\mathbb{I}_{\Nx}$. This is akin to a `\emph{smoothing}' operation of the target posterior. In this work, we use Annealed Langevin Dynamics, first proposed by \citet{song2019generative} and later improved in \cite{song2020improved}, to sample from the target posterior distributions using the learned score network starting from the smoothest target and gradually reducing the amount of smoothing; see \Cref{alg:ALD}. Line \ref{alg:line:langevin} in \Cref{alg:ALD} is similar to \Cref{eq:langevin-discrete} but uses the score network $\sm$ to approximate the score function $\nabla_{\x} \log \prob{\X \vert \Y}$ at different levels of smoothing. 
	\begin{algorithm}[t]
		\caption{Annealed Langevin dynamics~\cite{song2020improved}}\label{alg:ALD}
		\KwIn{Trained score network $\sm(\x, \y, \sigma;  \thetaa)$, measurement $\hat{\y}$, sampling steps $T$, step size parameter~$\epsilon$, and perturbation kernel standard deviations $\left\{ \sigma_i \right\}_{i=1}^L$}
		Initialize $\x_0$  such that its \supth{i} component $ \{ \x_0\}_i \sim \mathcal{U}(0, 1)$ \\
		\For{$i = 1$ \KwTo $L$}{
			$\alpha_i = \epsilon \sigma_i^2/\sigma^2_L$ \\
			\For{$j = 1$ \KwTo $T$}{
				Sample $\z \sim \mathcal{N}(\bm{0}, \Ind_{\Nx})$ \\
				$\x_j = \x_{j-1} + \alpha_i \sm(\x_{j-1}, \hat{\y}, \sigma_i; \thetaa) + \sqrt{2\alpha_i} \z$ \label{alg:line:langevin} \\
			}  
			Set $\x_0 = \x_T$ \\
		}
		Denoise $\x_\mathrm{final} = \x_0 + \sigma_L^2 s(\x_0, \hat{\y}, \sigma_L; \thetaa)$ \\
		\KwOut{Realization $\x_\mathrm{final}$}  
	\end{algorithm}

	\subsection{Hyper-parameters}\label{subsec:hyperparameters}
	
	There are a few hyper-parameters associated with cSDMs. Some of these hyper-parameters are related to the training of the score network while others are associated with the sampling algorithm used to generate new realizations. Here we discuss how to choose the various important hyper-parameters. 
	
	The hyper-parameters associated with score estimation include the noise levels $\sigma_1$ through $\sigma_L$, and the number $L$ of noise levels. Following \citet{song2020improved}, we choose $\sigma_1$ to be larger than the Euclidean distance between any two points of $\x$ within the training data. This helps smooth out the effect of $\prob{\X\vert\Y}$ from $\prob{\tilde{\X}\vert\Y}$ and sampling can commence from a near unimodal Gaussian distribution with covariance close to $\sigma_1^2 \mathbb{I}_{\Nx}$. This helps avoid realizations from getting stuck near modes of the target distribution and the sampling space can be properly explored with a limited number of realizations.  We choose $\sigma_L$ to be a very small value --- typically $0.001$ or $0.01$ --- so that the approximation $\prob{\tilde{\X}\vert\Y}(\tilde{\x} \vert \y)  \approx \prob{\X\vert\Y}(\x\vert \y) $ is better. The approximation improves as $\sigma_L$ decreases because $\lim_{\sigma_L \to 0} \prob{\tilde{\X}_{\sigma_L}\vert\Y}(\tilde{\x}_{\sigma_L} \vert \y) = \prob{\X\vert\Y}(\tilde{\x}_{\sigma_L} \vert \y)$ as $\probb{\tilde{\X}\vert\X}{\tilde{\x}\vert\x} $ converges to the Dirac delta function with $\sigma_L \to 0$.
	
	Following \citet{song2020improved}, we choose $\sigma_j$ to be a geometric progression between $\sigma_1$ and $\sigma_L$, \ie  we fix $\gamma = \sigma_{j-1}/\sigma_j$. Moreover, we choose the number $L$ of noise levels such that realizations from $\prob{\tilde{\X}_{\sigma_{j-1}}\vert\Y}$ are good initializers for the Langevin dynamics MCMC when exploring $\prob{\tilde{\X}_{\sigma_{j}}\vert\Y}$, which will be true when there is significant overlap between the two distribution. \citet{song2020improved} suggest choosing $\gamma$ such that:
	\begin{equation}\label{eq:chooseL}
		\Phi(\sqrt{2\Nx}(\gamma -1) + 3\gamma) - 	\Phi(\sqrt{2\Nx}(\gamma -1) - 3\gamma) \approx 0.5,
	\end{equation}
	where $\Phi$ is the cumulative distribution function of a standard normal variable, to ensure good overlap between successive smoothened versions of the posterior. In practice, we first choose $L$, compute $\gamma$ and then verify that \Cref{eq:chooseL} is satisfied. 
	
	Apart from the aforementioned hyper-parameters, there are the usual learning related hyper-parameters related to the training of the score network, which includes the learning rate, batch size and number of training iterations. These hyper-parameters are chosen to avoid over-fitting by monitoring the loss on a test set, which can be carved out from the initial training set if necessary.
	
	The hyper-parameters for the Annealed Langevin dynamics include the number of Langevin steps $T$ and the step size parameter $\epsilon$. Increasing $L$ will increase the total number of Langevin steps, and the total time it takes to generate realizations from the target posterior distribution. So, we choose $T$ as large as possible as the compute budget allows. Finally, we choose $\epsilon$ such that realizations from $\prob{\tilde{\X}_{\sigma_{j-1}}\vert\X}$ can reach their steady state distribution $\prob{\tilde{\X}_{\sigma_{j}}\vert\X}$ within $T$ steps. \citet{song2020improved} suggest choosing a value of $\epsilon$ for which~\cite{song2020improved}
	\begin{equation}\label{eq:choose_epsilon}
		\left( 1 - \frac{\epsilon}{\sigma^2_L}\right)^{2T} \left( \gamma^2 - \frac{2\epsilon}{\sigma^2_L - \sigma^2_L \left( 1 - \frac{\epsilon}{\sigma_L^2} \right)} \right) + \frac{2\epsilon}{\sigma_L^2 - \sigma^2_L\left( 1 - \frac{\epsilon}{\sigma_L^2} \right)} \approx 1. 
	\end{equation}
	In practice, we perform a grid search over probable values of $\epsilon$ and choose the value for which the left hand side of \Cref{eq:choose_epsilon} is closest to 1. 
	
	\subsection{Design choices for the score network}\label{subsec:design-score}
	
	The design choices for the score networks we use in this work follow suggestions made by \citet{song2020improved}. Primarily, the score network treats $\sigma_j$ unlike an usual input such as $\x$ or $\y$. Rather, $\sigma_j$ is used to scale the outputs from the network \ie $\sigma_j$ divides the output from the last layer of the score network. Moreover, we treat $\x$ and $\y$ as images as they are discretized forms of continuous field quantities: $\x$ is the spatially varying material constitutive parameter over the specimen discretized over an appropriate Cartesian grid, and $\y$ is the measured mechanical response of the specimen over the same Cartesian grid. Therefore, we use convolution neural network type architectures to construct the score functions with $\x$ and $\y$ as separate channels. More specifically, we use RefineNets~\cite{lin2017refinenet} to construct the score networks in this work. If there are multiple measurement types available or the measurement type itself has more than one component, we simply add additional channels. For instance, we treat vertical and horizontal displacement measurements as separate channels. Similarly, in the case of a time-dependent problem, if vertical displacement measurements are made at different time instants, the instantaneous measurements are concatenated as additional channels.

	\subsection{Workflow}\label{subsec:workflow}
	
\begin{figure}
		\centering
		\includegraphics[height=0.88\textheight]{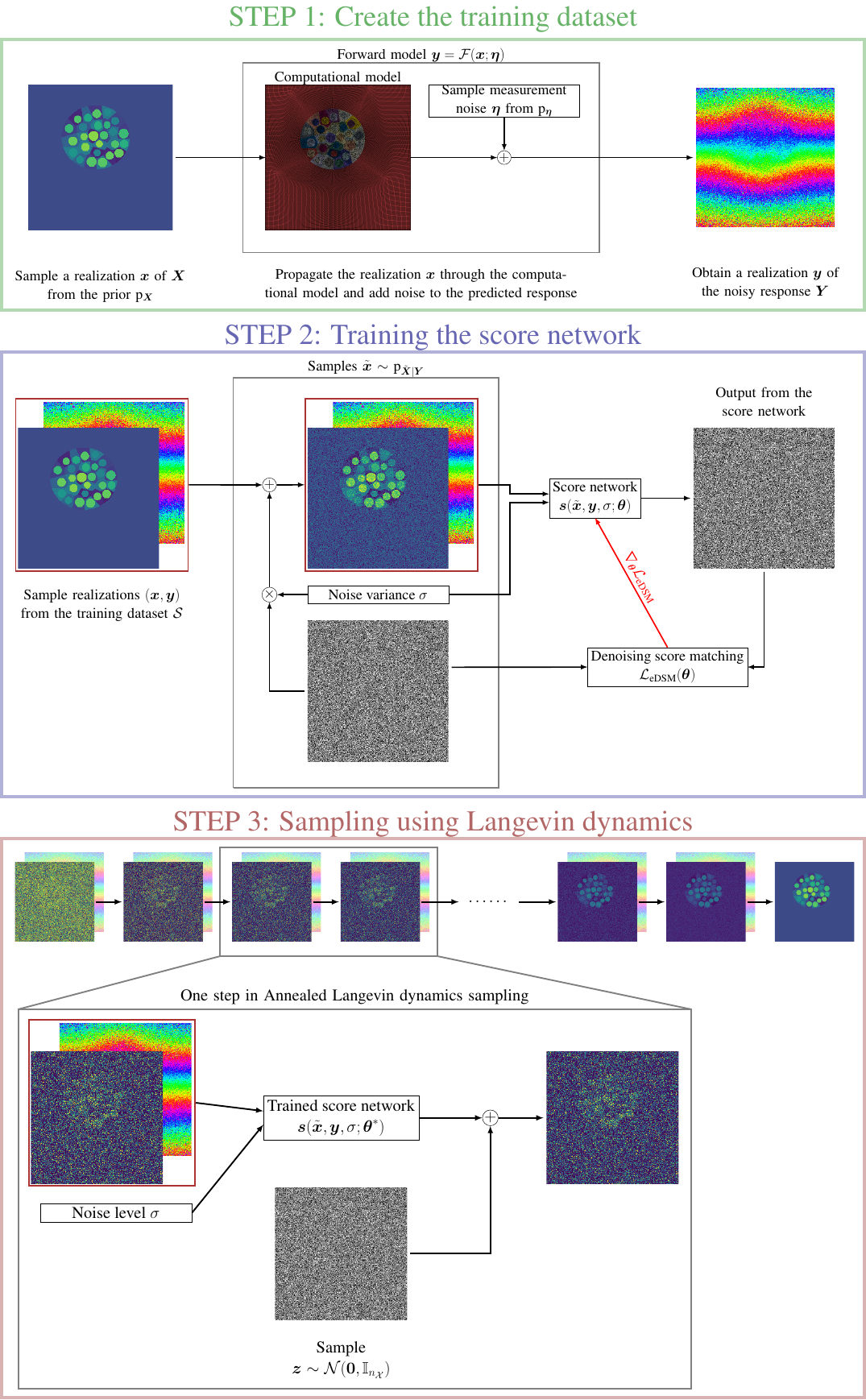}
		\caption{Schematic diagram of the proposed framework for solving inverse problems in mechanics using cSDMs. Step 1 involves creating the dataset by sampling realizations of $(\X, \Y)$ from the joint distribution $\prob{\X\Y}$, which requires multiple evaluations of the forward model $\mathcal{F}$. Step 2 involves the training of the score network, which progressively learns to predict the noise in realizations of $\tilde{\X}$ conditioned on $\Y$ for all levels of noise $\sigma$. Finally, Annealed Langevin dynamics is used to sample from the target posterior distribution in Step 3}
		\label{fig:workflow}
	\end{figure}

	In this section, we describe the workflow for solving an inverse problem using cSDMs. The basic workflow can be divided into three steps; see \Cref{fig:workflow}. 
	
	\subsubsection{Step 1: Create the training dataset}
	
	The first step involves the creation of the training dataset. For this step, we first develop a parameteric prior model for $\X$, resembling the first step in a more traditional application of Bayesian inference. Alternatively, multiple instances of $\X$ would also suffice. Next, for each realization $\x$ of $\X$ we solve the forward problem \Cref{eq:forward-model} to obtain a corresponding realization of $\y$ of $\Y$. Note that, to solve the forward model \Cref{eq:forward-model} we require access to the distribution of the noise $\bm{\eta}$ in the measurement model. Any other modeling errors must also be accounted for by \Cref{eq:forward-model}. Also note that simulating \Cref{eq:forward-model} will require simulating the underlying physics-based mechanics solver, which can be black-box and incompatible with auto-differentiation.  We denote using $\mathcal{S}$ the training dataset consisting $N_{\mathrm{train}}$ realizations of $\x$ and $\y$, \ie $\mathcal{S} = \left\{ \x^{(i)}, \y^{(i)}\right\}_{i=1}^{N_{\mathrm{train}}}$. 
	
	\begin{rem}
		We assume that we have access to a sufficient number of realizations of $\X$ to build the training dataset $\mathcal{S}$. This may be data that was previously acquired, or the result of sampling from a parametric prior developed to capture the uncertainty in $\X$. We also assume that we can sample from the distribution for $\bm{\eta}$ to account for the measurement noise or modeling errors in the prediction $\Y$. 
	\end{rem}
	
	\subsubsection{Step 2: Train the score network using the training dataset}
	
	After creating the training dataset $\mathcal{S}$, we train the score network. First, we determine the following hyper-parameters--- $\sigma_1$, $\sigma_L$ and $L$. Then we train the score network using a suitable optimization algorithm and the training objective \Cref{eq:eDSM-fianl}. As is typical in machine learning, at each iteration we do not estimate  $\mathcal{L}_{\mathrm{eDSM}}(\thetaa) $ using the entire training dataset, rather we use a small batch of training data.  Thus,  $\mathcal{L}_{\mathrm{eDSM}}(\thetaa) $ is estimated using $N_{\mathrm{batch}}$ training data points instead of $N_{\mathrm{train}}$ in \Cref{eq:eDSM-fianl}, where $N_{\mathrm{batch}}$ is the batch size. Additionally, the exponential moving averages of the weights are maintained. Let $\thetaa^{(i)}$ denote the parameters after the \supth{i} iteration and  $\thetaa^\prime$ denote an independent copy of those parameters. After every iteration, we update $\thetaa^\prime$ as:
	\begin{equation}
		\thetaa^\prime = m \thetaa^\prime + (1 - m) \thetaa^{(i)},
	\end{equation}
	where $m=0.999$ is a momentum parameter. This helps suppress any sudden fluctuations in the training loss, as is typically observed when training score networks~\cite{song2020improved}. At the end of training, we use the averaged parameters as the optimal parameters for the score network. 
	
	\subsubsection{Step 3: Generate samples from the posterior using Annealed Langevin dynamics}
	
	In this step, we sample from the target posterior via the Annealed Langevin dynamics algorithm shown in \Cref{alg:ALD}. First, we specify the number of Langevin steps $T$ and step size parameter $\epsilon$, and use the trained score network $\sm(\cdot, \cdot, \cdot; \thetaa^\prime)$ to generate new realizations of $\X$ for a given measurement $\hat{\y}$. \Cref{alg:ALD} is parallelizable and multiple realizations can be generated together. In practice, and for the numerical experiments in this paper, posterior realizations are generated in batches and, hence, in parallel. The batch size for sampling is not the same as, and often greater than, the batch size for training. The dimensionality of the inverse problem at hand and the memory capacity of the compute resource, CPU or GPU, dictates the sampling batch size. We use the generated realizations to estimate the posterior mean and posterior standard deviation of $\X$, with the former serving as the reconstruction and the latter providing a measure of uncertainty. 
	
	\begin{rem}\label{rem:score_network_approx}
				cSDMs only require samples from the joint distribution $\prob{\X \Y}$ between $\X$ and $\Y$ to train the conditional score network. Hence, the proposed approach is effectively a \emph{likelihood-free simulation-based} inference approach~\cite{cranmer2020frontier}. It is well known that the sampling efficiency of many traditional likelihood-free approaches scales poorly with the dimensionality of the inverse problem~\cite{cranmer2020frontier}. In contrast, the proposed approach scales with the dimensionality of the inverse problem at hand due to the neural network-based score function approximation. This scalability stems from the adeptness of the neural network in handling high-dimensional data and modeling complex patterns between them. Moreover, the trained conditional score network approximates the posterior's score function $\nabla_{\x} \log \probb{\X \vert \Y}{\x \vert \hat{\y}}$ for all realizations of $\hat{\y}$. Therefore, using a trained score network, we can sample the posterior distributions for different realizations of $\Y$ without re-training the score network. This helps `\emph{amortize}' the inference cost over $\Y$ \cite{baptista2024conditional} and makes the proposed approach particularly suitable for high throughput applications requiring repeated solving of the same inverse problem. A comparative study of cSDMs with other simulation-based and likelihood-free inference methods is beyond the scope of the current work.
	\end{rem}
	
	\section{Results}\label{sec:results}
	
	In this section, we use cSDMs to solve various large-scale inverse problems in mechanics where the goal is to infer the Young's modulus (or a related constitutive parameter) field of a material specimen subject to external loading. We consider the following problems:
	\begin{enumerate}[itemsep=0pt]
		\item \Cref{subsubsec:elasto} --- Quasi-static elastography problem of inferring Young's modulus field of specimens containing a stiff inclusion within a soft homogeneous background media from displacement measurements.
		\item \Cref{subsubsec:helmholtz} --- Time-harmonic elastography problem of inferring the shear modulus field of regions surrounding the human optic nerve head (ONH) from complex measurements of a wave field as the specimen is subject to harmonic excitation.
		\item \Cref{subsubsec:elasto2} --- Quasi-static elastography problem similar to \Cref{subsubsec:elasto} but using experimental data from tissue-mimicking hydrogels~\cite{pavan2012nonlinear}. 
		\item \Cref{subsubsec:helmholtz2} --- Time-dependent elastography problem of inferring Young's modulus of a dual layer tissue-mimicking phantom excited by acoustic radiation force using snapshots of the displacement field taken at regular time intervals from a laboratory experiment~\cite{lu2021layer}. 
		\item \Cref{subsubsec:TS} --- Optical coherence elastography (OCE) of tumor spheroids (clusters of cells within a hydrogel) wherein the Young's modulus field of the specimen is inferred from the phase difference between optical coherence microscopy scans as the specimen undergoes compression. This example also involves experimental data and its details can be found in \citet{foo2024tumor}. 
	\end{enumerate}
	\begin{table}[t]
		\small
		\centering
		\caption{Summary of inverse problems considered in this work}
		\label{tab:problem_details}
		\begin{tabular}{@{\extracolsep{2pt}} lc cccc}
			\toprule[1.5pt]
			& \multirow{2}{*}{Inverse problem} & \multirow{2}{*}{\makecell{Forward\\physics}} & \multicolumn{2}{c}{Dimensionality} & \multirow{2}{*}{\makecell{Training dataset\\size $N_{\mathrm{train}}$}} \\
			\cline{4-5}
			& & & $\Nx$ & $\Ny$&\\	
			\midrule[1pt]
			\multirow{2}{*}{\rotatebox{90}{\makecell{Synthetic\\data}}}& \makecell{Circular inclusion\\ (\Cref{subsubsec:elasto})} & \makecell{Linear,\\quasi-static} & 56$\times$56 & 56$\times$56  & 10,000\\
			& \makecell{Optic nerve head\\ (\Cref{subsubsec:helmholtz})} & \makecell{Linear,\\time-harmonic} & 64$\times$64 & 2$\times$64$\times$64 & 12,000\\
			\midrule[0.5pt]
			\multirow{3}{*}{\rotatebox{90}{\makecell{Experimental\\data}}}& \makecell{Circular inclusion\\ (\Cref{subsubsec:elasto2})} & \makecell{Linear,\\quasi-static} & 56$\times$56 & 56$\times$56  & 10,000\\
			& \makecell{Dual-layer phantom\\ (\Cref{subsubsec:helmholtz2})} & \makecell{Linear,\\time-dependent}& 64$\times$64 & 10$\times$64$\times$64 & ~3,000\\
			& \makecell{Tumor spheroids\\ (\Cref{subsubsec:TS})} & \makecell{Linear,\\quasi-static} & 256$\times$256 & 256$\times$256  & 24,000\\
			\bottomrule[1.5pt]
		\end{tabular}
	\end{table}	
	\Cref{tab:problem_details} provides an overview of the various inverse problems we tackle including the type of measurements, the nature of the forward models, the dimensionality $\Nx$ and $\Ny$ of the inferred quantity and measurements, respectively, and the corresponding size of the training dataset used to train the score network. More specifically, the numerical examples have the following variety:
	\begin{enumerate}
		\item \emph{Synthetic and real measurements} --- In \Cref{subsec:synthetic}, we consider two inverse problems with synthetic measurements. In particular, the quasi-static elastography inverse problem in \Cref{subsubsec:elasto} has a relatively simple latent prior distribution amenable to Monte Carlo simulation (MCS), which provides an opportunity to validate the proposed method.
		In \Cref{subsec:experimental}, we apply cSDMs to solve inverse problems with measurements coming from physical experiments. However, note that even in such cases, we train the score network using synthetic data generated from the joint distribution that we model. 
		\item \emph{Forward physics} --- In general, the governing mechanics model are the following equations of equilibrium and linear elastic constitutive law
		\begin{gather}
			\nabla \cdot \bm{\sigma} + \rho \bmm = \frac{\partial^2\uu}{\partial t^2} , \label{eq:elastodynamic_equation} \\
			\bm{\sigma}= 2\mu \nabla^{\mathrm{s}}\uu + \lambda (\nabla\cdot\uu) \bm{I} , \label{eq:linear_constitutive_law}
		\end{gather}
		for an isotropic solid, along with appropriate boundary conditions that we specify later.  In \Cref{eq:elastodynamic_equation,eq:linear_constitutive_law}, $\bm{\sigma}$ is the Cauchy stress tensor, $\rho$ denotes the material density, $\bmm$ denotes the body forces, $\uu$ denotes the displacement field, $\nabla^{\mathrm{s}}\uu$ denotes the strain tensor, and $\lambda$ and $\mu$ are the first and second Lam\'{e} parameters, respectively. For the inverse problems in \Cref{subsubsec:elasto,subsubsec:elasto2,subsubsec:TS}, which involve quasi-static conditions, \Cref{eq:elastodynamic_equation} simplifies to 
		\begin{equation}
			\nabla \cdot \bm{\sigma} = 0, \label{eq:equilibrium_equation}
		\end{equation}
		in the absence of body forces. However, the inverse problem in \Cref{subsubsec:helmholtz2} requires the equilibrium equation \Cref{eq:elastodynamic_equation} to account for inertia and time-dependent body forces. For the inverse problem in \Cref{subsubsec:helmholtz}, the forward physics relating the complex-valued displacement field with the material shear modulus field, $\mu$, is modeled using the Helmholtz equation:
		\begin{equation}\label{eq:helmholtz}
			- \omega^2 \left(  u_{\mathrm{R}} + i u_{\mathrm{I}}   \right) - \nabla \cdot \left[ \frac{\mu}{\rho} \left( 1 + i \alpha \omega \right) \nabla \cdot \left(  u_{\mathrm{R}} + i u_{\mathrm{I}}   \right)  \right] = 0 ,
		\end{equation}
		where $\omega$ is the frequency of excitation, $u_{\mathrm{R}}$ and $u_{\mathrm{I}}$ denote the real and imaginary components of the displacement, $\mu$ denotes the shear modulus field, $\alpha$ is the wave dissipation coefficient, and $\rho$ is the material density. 
		\item \emph{Measurement noise and modeling errors} --- In \Cref{subsubsec:elasto,subsubsec:helmholtz,subsubsec:elasto2}, we model the measurement noise as homoscedastic zero-mean Gaussian random variables. The measurement noise is non-Gaussian and non-additive in the inverse problems that we consider in \Cref{subsubsec:helmholtz2,subsubsec:TS}. This shows that the proposed method can handle complex measurement noise models. Additionally, the forward model in \Cref{subsubsec:helmholtz2} contains random variables that account for the modeling error, which further highlights the flexibility of the proposed method in accommodating complex forward models. 
		\item \emph{Black-box forward models} --- In addition to the complex noise, we use finite element models to simulate the response of a specimen. These models are implemented using suitable finite element packages that are used as a black-box.
		\item \emph{Types of measurements} --- For the inverse problems in \Cref{subsubsec:elasto,subsubsec:elasto2,subsubsec:TS}, the underlying physics is quasi-static and we consider only the vertical deformations. Therefore, the measurements comprise an instantaneous displacement field after deformation (as in \Cref{subsubsec:elasto,subsubsec:elasto2}) or a quantity derived from the relative deformations of the specimen prior to and after undergoing deformation (as in \Cref{subsubsec:TS}). The measurements are akin to an image with a single channel in these cases. For the inverse problem in \Cref{subsubsec:helmholtz}, the measurement is a discretized complex-valued displacement field, and its real and imaginary components are treated as two separate channels of an image. Similarly, for the inverse problem in \Cref{subsubsec:helmholtz2}, each instantaneous displacement field is treated as a separate channel, leading to a total of 10 channels. 
		
		\item \emph{High dimensionality} --- All the inverse problems involve the inference of a high-dimensional quantity as a result of discretizing the Young's modulus field over a cartesian grid. In this regard, the inverse problem in \Cref{subsubsec:TS} where $\Nx = $256$\times$256 is the most challenging. 
	\end{enumerate}
	
	\paragraph{Details of implementation} We implement the proposed method on \texttt{PyTorch}~\cite{pytorch2019}. Before training the score network, we min-max normalize the training dataset to the range $[0,1]$. We perform this normalization independently for the realizations of $\X$ and $\Y$ in the training dataset. For the synthetic problems in \Cref{subsec:synthetic}, we present all results in normalized units,  meaning all posterior statistics are also estimated in the normalized units. For the inverse problems where we have experimental measurements, we present all results in physical units after inverting the normalization. Additionally, for the tumor spheroid application in \Cref{subsubsec:TS}, we take the logarithm of the Young's modulus field relative to the Young's modulus of the surrounding media (hydrogel) before normalizing the realizations of $\X$ in the training dataset. This helps improve the scaling of the training data because the Young's modulus may vary by many orders of magnitude between the cells within a tumor spheroid and the surrounding media. We list the training and sampling hyper-parameters for the various inverse problems in~\Cref{tab:hyper-parameters}. 
	
	\begin{table}[t]
		\centering
		\caption{Training and sampling hyper-parameters for the cSDMs}
		\label{tab:hyper-parameters}
		{\small
			\begin{tabular}{@{\extracolsep{1pt}} l cc ccc}
				\toprule[1.5pt]
				\multirow{2}{*}{Hyper-parameters}& \multicolumn{5}{c}{Inverse problem} \\
				\cline{2-6}
				& \multicolumn{2}{c}{Synthetic data}  & \multicolumn{3}{c}{Experimental data} \\
				\cline{2-3} \cline{4-6}
				& \makecell{Circular\\inclusion\\ (\S~\labelcref{subsubsec:elasto})} & \makecell{Optic nerve\\head \\ (\S~\labelcref{subsubsec:helmholtz})} & \makecell{Circular\\inclusion\\ (\S~\labelcref{subsubsec:elasto2})} & \makecell{Dual-layer\\phantom\\ (\S~\labelcref{subsubsec:helmholtz2})} & \makecell{Tumor\\spheroids \\ (\S~\labelcref{subsubsec:TS})} \\
				\midrule[1pt]
				First noise level $\sigma_1$ & 20 & 50& 50& 50 & 120 \\
				Final noise level $\sigma_L$ & 0.01 & 0.01& 0.01& 0.01 & 0.01 \\
				No. of noise levels $L$ & 128 & 256& 256& 256 & 680 \\
				No. of Langevin steps $T$ & 5 & 5& 5& 5 & 5 \\
				Step size parameter $\epsilon$ & \pwr{5.7}{$-$6} & \pwr{5.7}{$-$6}& \pwr{5.7}{$-$6} & \pwr{5.7}{$-$6} & \pwr{5.7}{$-$6} \\
				Learning rate& 0.0001 & 0.0001 & 0.0001 & 0.0001 & 0.0001 \\
				Batch size & 128 & 64 & 128& 128 & 32 \\
				Total training epochs & 100,000 & 300,000 & 400,000 & 300,000 & 300,000 \\
				\bottomrule[1.5pt]
		\end{tabular}}
	\end{table}	
	\begin{table}[H]
		\small
		\centering
		\caption{Approximate computational cost of using cSDMs for Bayesian inference}
		\label{tab:compute_cost_breakdown}
		\begin{threeparttable}
			\begin{tabular}{@{\extracolsep{2pt}} l cc cc cc}
				\toprule[1.5pt]
				\multirow{2}{*}{\makecell{Inverse\\problem}} & \multicolumn{2}{c}{Step 1\textsuperscript{$\ast$}} & \multicolumn{2}{c}{Step 2\textsuperscript{$\dagger$}} & \multicolumn{2}{c}{Step 3\textsuperscript{$\ddagger$}} \\
				\cline{2-3} \cline{4-5} \cline{6-7}
				& Wall time& Resource& Wall time& Resource& Wall time& Resource\\
				\midrule
				\makecell{Circular inclusion\\ (\Cref{subsubsec:elasto})}  & 11 s & \makecell{4 $\times$ 12 core\\Intel Xeon\\4116 CPU}& 43 s & \makecell{NVIDIA\\A40 GPU} & 2 mins & \makecell{2 $\times$ NVIDIA\\P100 GPU} \\
				\makecell{Optic nerve head\\ (\Cref{subsubsec:helmholtz})} & 17 min & \makecell{8 core\\Apple\\M1 Pro} & 30 s & \makecell{NVIDIA\\A40 GPU} & 2 mins & \makecell{NVIDIA\\A40 GPU} \\
				\makecell{Circular inclusion\\ (\Cref{subsubsec:elasto2})} & 11 s & \makecell{4 $\times$ 12 core\\Intel Xeon\\4116 CPU} & 1 min & \makecell{NVIDIA\\Quadro\\RTX 8000} & 3 mins & \makecell{NVIDIA\\Quadro\\RTX 8000} \\
				\makecell{Dual-layer phantom\\  (\Cref{subsubsec:helmholtz2})} & 1.5 hrs & \makecell{4 $\times$ 4 core\\Intel i5\\7400 CPU} & 35 s & \makecell{NVIDIA \\ A100 GPU}  & 90 s & \makecell{NVIDIA \\ A100 GPU} \\
				\makecell{Tumor spheroids\\ (\Cref{subsubsec:TS})} & 2.5 hrs\textsuperscript{$\star$}  &\makecell{2 $\times$ 8 core\\Intel Xeon\\E5-2690 CPU}& 2 mins &  \makecell{2 $\times$ NVIDIA\\A40 GPU}& 45 mins & \makecell{2 $\times$ NVIDIA\\A40 GPU}\\
				\bottomrule[1.5pt]
			\end{tabular}
			\begin{tablenotes}
				\item \textsuperscript{$\ast$}For generating 100 training data points.
				\item  \textsuperscript{$\dagger$}For 100 training iterations.
				\item \textsuperscript{$\ddagger$}For sampling 100 posterior realizations.
				\item  \textsuperscript{$\star$}Reported in a previous study by \citet{foo2024tumor}
			\end{tablenotes}
		\end{threeparttable}
	\end{table}
	
	\paragraph{Breakdown of computational costs for various steps in the workflow} \Cref{tab:compute_cost_breakdown} provides an estimate of the wall times (a proxy for the computational cost) for the three steps in the workflow and the resource type used. In general, we make the following remarks. The cost of generating the training dataset depends on the complexity and dimensionality of the problem. The cost of training the score network depends on the batch size and the inverse problem’s dimensionality. The time required for posterior sampling is influenced by the total number of Langevin steps, which equals $L \times T$. This is perhaps the main drawback of using diffusion models compared to other generative models and remains an active research area.

	\subsection{Experiments on synthetic data}\label{subsec:synthetic}
	
	In this section, we apply cSDMs to two inverse problems with synthetic data. The inverse problem in \Cref{subsubsec:elasto} helps validate the proposed approach by comparing the solutions with MCS.
	
	\subsubsection{Quasi-static elastography}\label{subsubsec:elasto}
	
\begin{figure}[t]
		\centering
		\includegraphics[width=\textwidth]{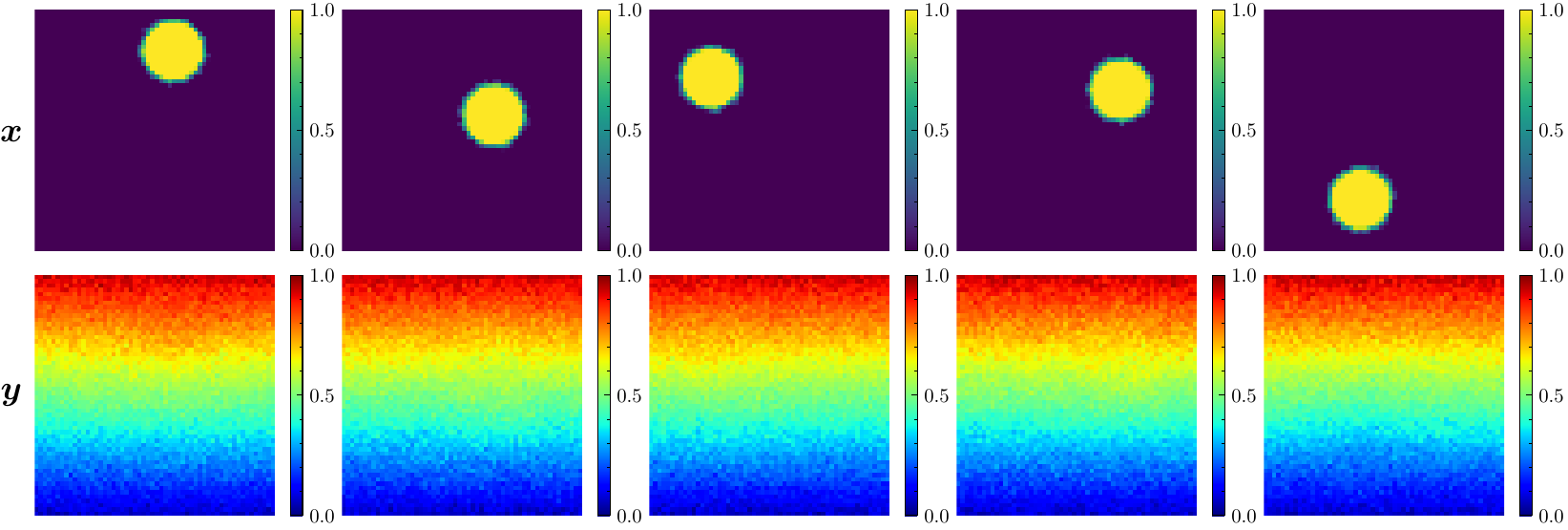}
		\caption{Five typical realizations of $\X$ and $\Y$ sampled from the joint distribution for the synthetic quasi-static elastography example and belong to the training dataset. The first row shows the shear modulus field, while the second row shows the corresponding noisy vertical displacement field used as measurements. In this figure, the standard deviation of the Gaussian noise $\sigma_\eta = 0.025$. Moreover, realizations of $\X$ are $\Y$ are min-max normalized to $[0,1]$ scale}
		\label{fig:sci_sample}
	\end{figure}

	The first synthetic example concerns quasi-static elastography, a medical imaging technique in which tissues undergoing deformations under an external load are imaged using ultrasound~\cite{barbone2007elastic,barbone2010review}. The ultrasound images are used to determine the displacements of the heterogeneous specimen, and the goal is to infer the spatially varying shear modulus. This study features a tissue phantom embedded with a stiffer circular inclusion within a softer surrounding matrix, such as the realizations shown in the first row of \Cref{fig:sci_sample}. In elastography applications, ultrasound can accurately measure the displacement component along the transducer axis, which in this case is the vertical direction. Thus, the inverse problem comprises inferring the spatial distribution of the shear modulus from noisy measurements of vertical displacements of the phantom. The mechanics is governed by the linear elastic model \Cref{eq:equilibrium_equation} and \labelcref{eq:linear_constitutive_law}. Moreover, we assume that the specimen is incompressible and in plane stress condition following previous studies~\cite{patel2021solution,ray2023solution}.
	
	In this example, the dimensions of the specimen is 1$\times$1~cm\textsuperscript{2}. To simulate compression, we fix the bottom edge and impose a downwards vertical displacement of 0.01 cm on the top edge. The bottom edge is traction-free in the horizontal direction, whereas the left and right edges are kept traction-free in both directions. Additionally, we pin the bottom left corner to prevent rigid body motion. A realization of $\bm{X}$ corresponds to the spatial distribution map of the shear modulus of the specimen discretized over a 56$\times$56 Cartesian grid. We formulate a parametric prior that represents a uniformly stiff circular inclusion set against a homogeneous background: the coordinates of the inclusion's center are uniformly distributed over the domain such that the inclusion fits within the boundaries. We intentionally work with a simple prior, consisting of only two random variables, to facilitate MCS. See \Cref{tab:sci_prior} for details regarding the prior distribution for this example. We fix the inclusion's radius to be 0.12~cm and choose the shear modulus of the inclusion and the background to be 1.5 and 0.1 kPa, respectively. 
	\begin{table}[t]
		\caption{Random variables comprising the parametric prior for the synthetic quasi-static elastography problem. Note, $\mathcal{U}(a,b)$ denotes an uniform random variable between $a$ and $b$}
		\label{tab:sci_prior}
		\centering
		\begin{tabular}{lcc}
			\toprule
			Random variable & Distribution  \\
			\toprule
			Distance of the inclusion's center from the left edge (mm) & $\mathcal{U}(0.2,0.8)$ \\
			Distance of the inclusion's center from the bottom edge (mm) & $\mathcal{U}(0.2,0.8)$ \\
			\bottomrule
		\end{tabular}
	\end{table} 
	
	The training dataset comprises 10,000 pairwise realizations of $\X$ and $\Y$. For a given realization $\x$ of $\X$, we generate the corresponding measurement $\y$, which is a realization of $\Y$, by utilizing $\x$ in a finite element model consisting of linear triangular elements. We use FEniCS~\cite{LoggEtal2012} to conduct the finite element analysis. We discretize the measurement over the same 56$\times$56 Cartesian grid as the shear modulus field. Therefore, $\Nx = \Ny =$ 56$\times$56 in this example. Subsequently, we add measurement noise --- homoscedastic Gaussian noise with standard deviation $\sigma_\eta \times u_{\mathrm{max}}$ truncated to within the $[-3\sigma_\eta \times u_{\mathrm{max}}, 3\sigma_\eta \times u_{\mathrm{max}}]$ range, where $u_{\mathrm{max}}$ is the maximum vertical displacement across all specimens in the training dataset.  \Cref{fig:sci_sample} illustrates five randomly selected realizations from the training dataset. Note that in the training data, both $\X$ and $\Y$, are min-max normalized to the range $[0,1]$. We work with three values of $\sigma_\eta \times u_{\mathrm{max}}$ corresponding to varying intensity of measurement noise --- 2.5\%, 5\%, and 10\% of $u_{\mathrm{max}}$ --- to create three training data sets, which we then use to train the score network. Note, we train a separate score network for each level of measurement noise. \Cref{tab:hyper-parameters} lists the training and sampling hyper-parameters we use for this problem. After training the score network, we employ Annealed Langevin dynamics, as described in \Cref{subsec:sampling}, to sample the posterior distribution. 
	
\begin{figure}
		\begin{tikzpicture}
			\node[draw=gray!00,inner sep=2pt](A) at (0,0) {\includegraphics[width=0.99\textwidth]{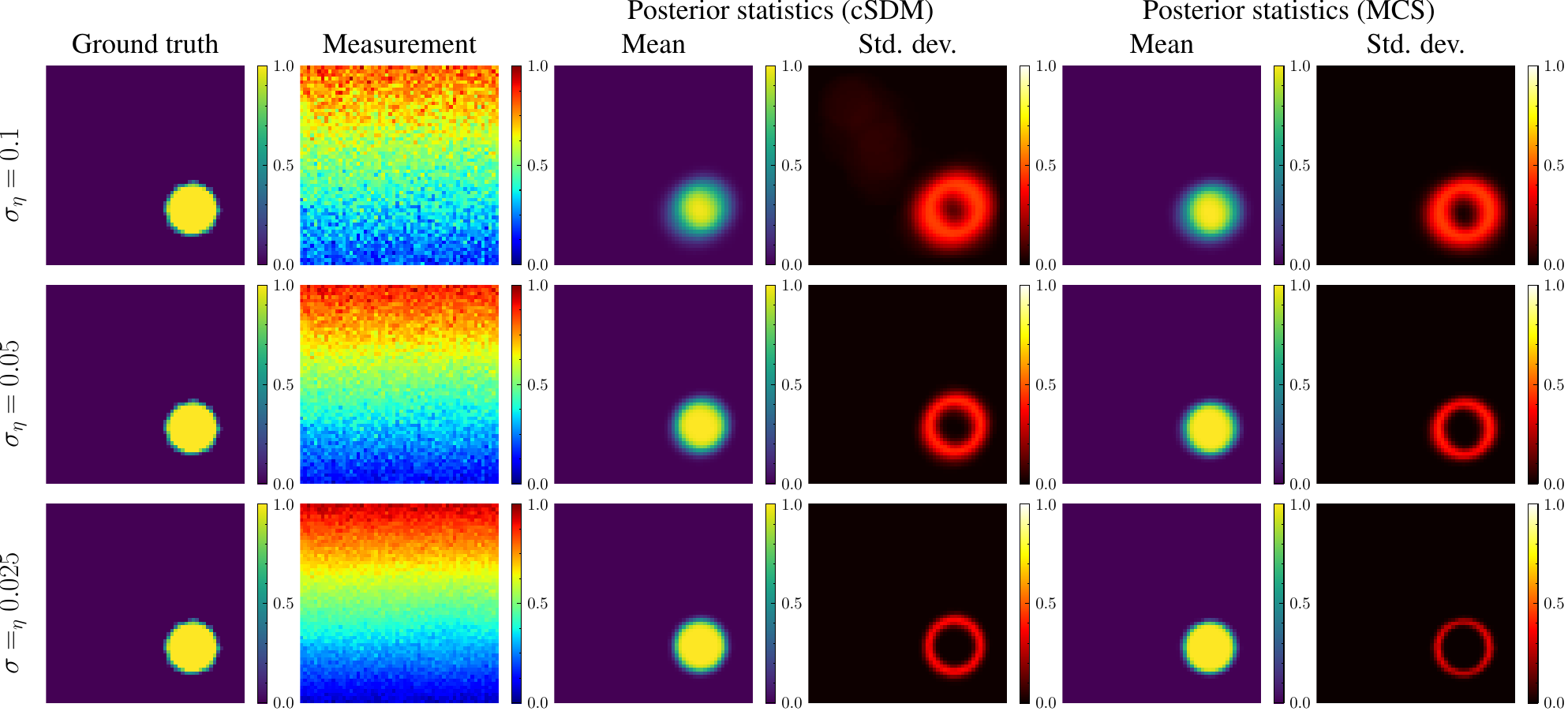}};
			\node[draw=gray!00,inner sep=2pt,anchor=north,yshift=-40pt](B) at (A.south) {\includegraphics[width=0.99\textwidth]{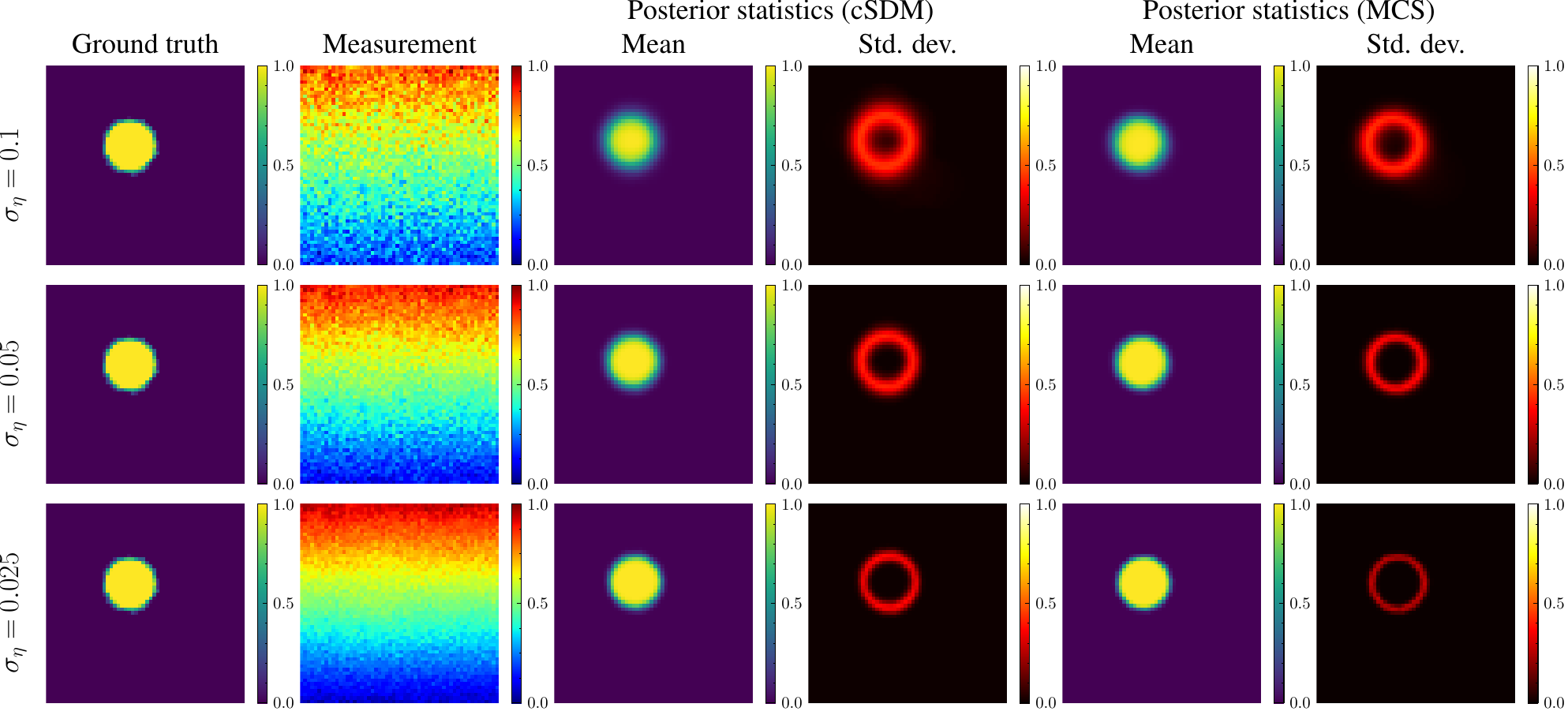}};
			
			\node[outer sep=2pt,anchor=north](A1) at (A.south) {(a) Test sample 1};
			\node[outer sep=2pt,anchor=north](B1) at (B.south) {(b) Test sample 2};     
		\end{tikzpicture}
		\caption{Posterior statistics -- pixel-wise posterior mean and standard deviation -- estimated using cSDMs and MCS for the synthetic quasi-static elastography example on two test samples. In subfigures (a)~and~(b), the three rows show the results corresponding to three different levels of measurement noise $\sigma_\eta$ but for the same ground truth. All values are normalized to $[0,1]$ scale}
		\label{fig:sci_test}
	\end{figure}

	We also estimate the posterior statistics using MCS, which we treat as the true statistics, to validate the results obtained using the cSDMs. We use a sample size equal to 500,000 for the MCS, which is significantly more than the size of the training dataset. Note that the measurement noise is additive in this case and the likelihood function can be used to obtain the importance weights, which we subsequently self-normalize. See \ref{app:MCS} for details of the MCS procedure. We compare the posterior statistics estimated using cSDM and MCS on two test samples that were not part of the training dataset; \Cref{fig:sci_test} shows the results. In \Cref{fig:sci_test}(a)~and~(b), column 1 shows the test phantom while column 2 shows the corresponding measurement for three different noise levels. Note that the ground truth is not readily discernible through visual inspection of the measurements in column 2 of \Cref{fig:sci_test}.
	\begin{table}[t]
		\caption{Root mean squared error (RMSE) between the pixel-wise posterior statistics estimated using cSDM and MCS, for the test phantoms in \Cref{fig:sci_test}}
		\label{tab:sci_poststatsRMSE}
		\centering
		\begin{tabular}{@{\extracolsep{5pt}} c cc cc}
			\toprule
			\multicolumn{5}{c}{RMSE in posterior statistics} \\
			\midrule
			\multirow{2}{*}{\makecell{Measurement\\noise ($\sigma_\eta$)}} & \multicolumn{2}{c}{Mean} & \multicolumn{2}{c}{Std. Dev.}\\
			\cline{2-3} \cline{4-5}
			& Test sample 1& Test sample 2 & Test sample 1 & Test sample 2 \\
			\midrule
			10.0\% & 0.036 & 0.027 & 0.038 & 0.029 \\
			~5.0\% &  0.039 & 0.025 & 0.032 & 0.029 \\
			~2.5\% & 0.031 & 0.022 & 0.028 & 0.027 \\
			\bottomrule
		\end{tabular}
	\end{table} 
	
	\begin{table}[t]
		\caption{Root mean squared error (RMSE) between the pixel-wise posterior statistics estimated using cSDM and MCS, averaged over ten test phantoms for the synthetic quasi-static elastography example}
		\label{tab:sci_poststatsAvgRMSE}
		\centering
		\begin{tabular}{ccc}
			\toprule
			\multicolumn{3}{c}{Average RMSE in posterior statistics} \\
			\midrule
			\makecell{Measurement\\noise ($\sigma_\eta$)}  & Mean & Std. Dev.\\
			\toprule
			10.0\% & 0.051 & 0.051 \\
			~5.0\% &  0.031 & 0.032\\
			2.5\% & 0.025 & 0.029\\
			\bottomrule
		\end{tabular}
	\end{table} 
	\Cref{fig:sci_test} also shows the pixel-wise posterior mean and standard deviation estimated using the cSDM and MCS. We generate 10000, 4000, and 1000 posterior realizations to compute these statistics using the trained score network for $\sigma_\eta =$10\%, 5\%, and 2.5\%, respectively. These sample sizes were roughly close to the effective sample size of the MC samples for the three cases. \Cref{fig:sci_test} shows that the posterior mean estimated using cSDM successfully locates the inclusion at all measurement noise levels, similar to MCS. Furthermore, the posterior statistics estimated using cSDM are qualitatively similar to those obtained using MCS, with the root mean squared errors (RMSE) listed in Table \ref{tab:sci_poststatsRMSE}. Both these methods predict larger uncertainty around the edges of the inclusion, and the uncertainty predictably grows with the level of measurement noise. For the lowest level of measurement noise, the reconstructed inclusion in the posterior mean is sharp as there is very little uncertainty, both in spread and magnitude, surrounding the inclusion. The cSDM yields results similar to MCS for other test phantoms as well. \Cref{tab:sci_poststatsAvgRMSE} tabulates the RMSE between the posterior statistics estimated using cSDM and MCS across ten test phantoms. The consistently low average RMSE values suggest that the proposed method is accurate and robust to the amount of measurement noise. 
	
\begin{figure}[h]
			\includegraphics[width=0.99\textwidth]{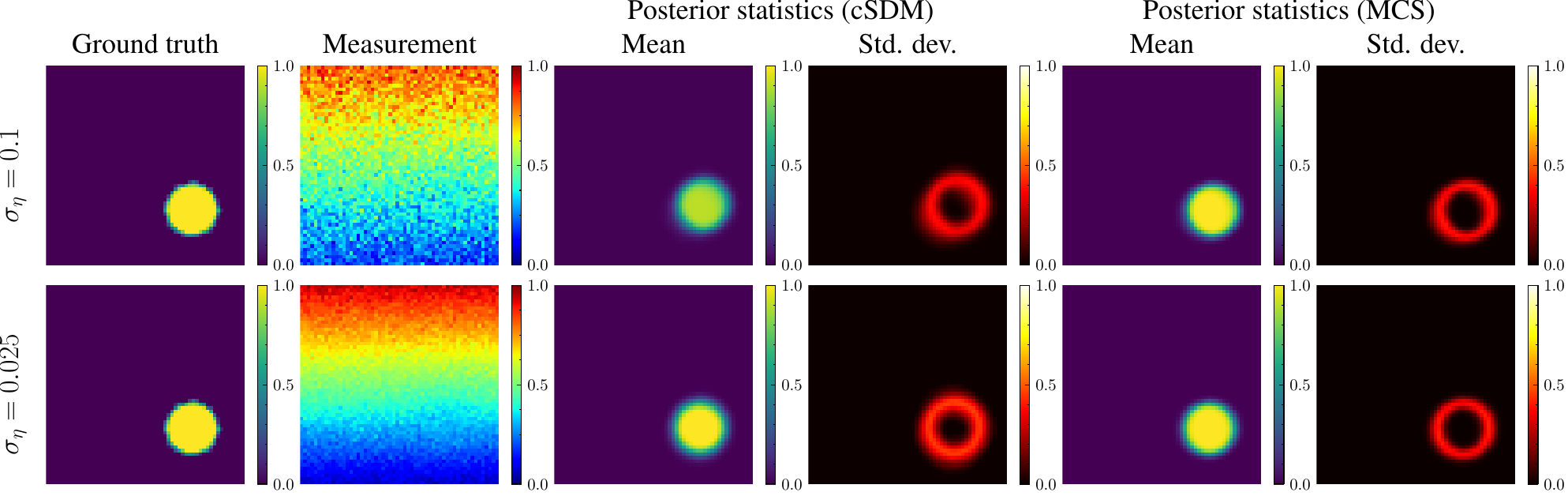}
			\caption{Posterior statistics -- pixel-wise posterior mean and standard deviation -- for test sample 1 estimated using cSDMs trained assuming measurement noise with standard deviation $\sigma_{\eta} = 0.05$. The reference MC statistics are also estimated with the misspecified measurement noise model. All values are normalized to $[0,1]$ scale}
			\label{fig:sci_incorrect_noise}
		\end{figure}

	\paragraph{Effect of model misspecfication} To understand the effects of a misspecified model for $\prob{\Y\vert\X}$, we perform inference for the case when the measurement noise's standard deviation $\sigma_{\eta} = 0.025$ and $0.1$ with the score network trained assuming $\sigma_{\eta}=0.05$. For test sample 1, \Cref{fig:sci_incorrect_noise} shows the posterior statistics estimated using the cSDM and compares it against the MCS under the misspecified noise model. Comparing \Cref{tab:sci_poststats_incorrectnoise,tab:sci_poststatsRMSE}, the RMSE in the posterior statistics estimated using the cSDM increases when the noise model is misspecified. In particular, \Cref{fig:sci_incorrect_noise} and \Cref{tab:sci_poststats_incorrectnoise} show that the pixel-wise posterior mean is particularly biased when the noise in the measurement has standard deviation $\sigma_{\eta}=0.1$ but inference is performed assuming $\sigma_{\eta}=0.05$. These results highlight the need for studying the effects of model misspecification on the performance of cSDMs. 
		\begin{table}[h]
			\caption{Root mean squared error (RMSE) between the pixel-wise posterior statistics estimated using cSDM and MCS for misspecified values of $\sigma_{\eta}$}
			\label{tab:sci_poststats_incorrectnoise}
			\centering
			\begin{tabular}{ccc}
				\toprule
				\makecell{Measurement\\noise ($\sigma_\eta$)}  & Mean & Std. Dev.\\
				\midrule
				10.0\% & 0.069 & 0.038 \\
				~2.5\% & 0.024 & 0.037 \\
				\bottomrule
			\end{tabular}
	\end{table}
	
	\subsubsection{Time-harmonic elastography}\label{subsubsec:helmholtz}
	
\begin{figure}[t]
		\centering
		\includegraphics[width=0.3\linewidth]{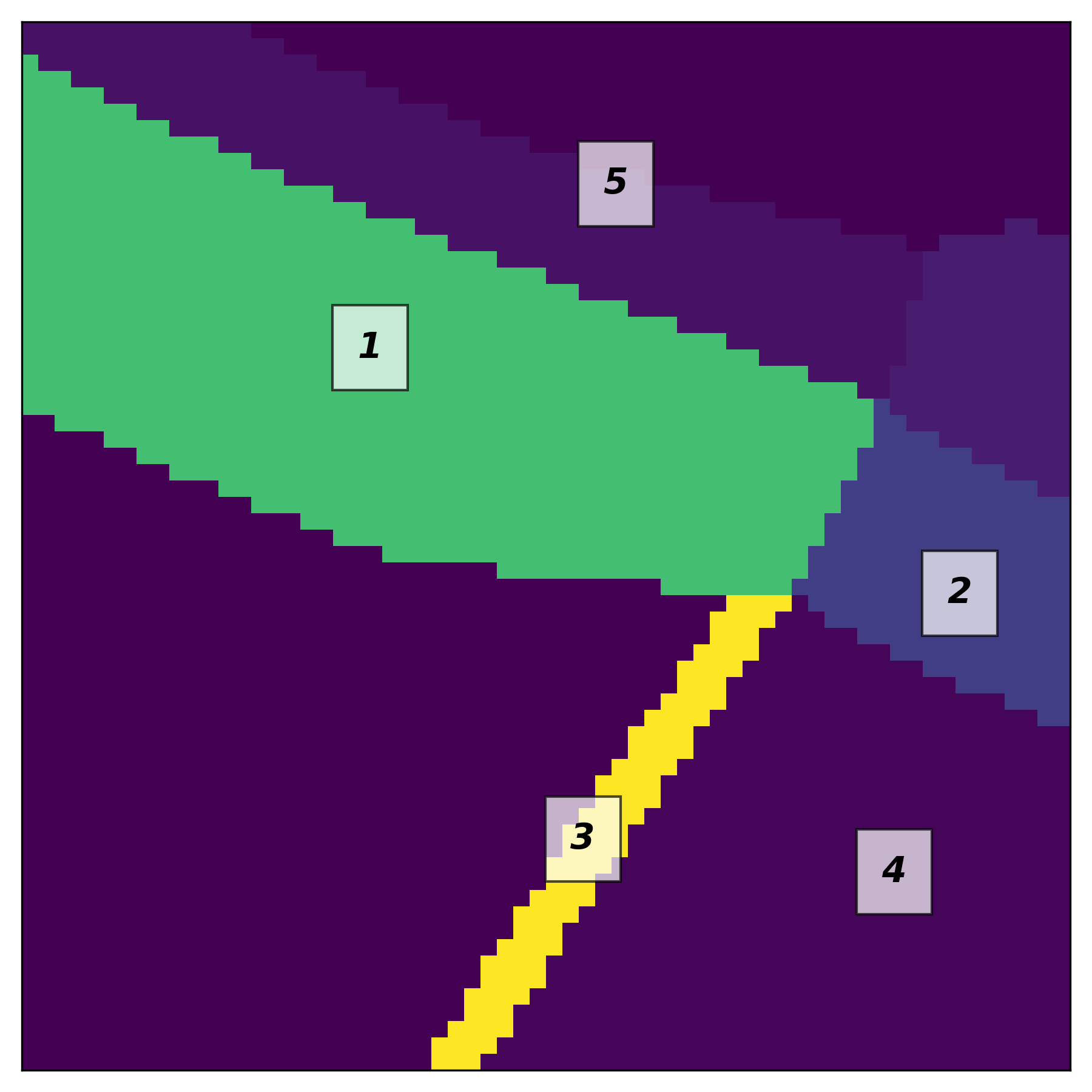}
		\caption{A realization of the ONH sampled from the parametric prior distribution showing the (1) peripapillary sclera, (2) lamina cribrosa, (3) pia matter, (4) optic nerve, and (5) retina. This figure also appears in~\cite{ray2023solution}}
		\label{fig:onh}
	\end{figure}

	In this example, we consider the inverse Helmholtz problem, which arises in elastography applications wherein the heterogeneous mechanical properties of a specimen must be recovered from measurements corresponding to time-harmonic excitation. More specifically, in this example, the shear modulus of the tissue surrounding a typical human ONH is to be inferred from measurements of propagating shear waves. \Cref{fig:onh} shows a typical realization from the prior and labels the various parts surrounding the ONH. We adapt this example from a previous study by \citet{ray2023solution}. 
	
	The forward model comprises of the Helmholtz equation, \Cref{eq:helmholtz}, where we choose $\alpha=$ \pwr{5}{$-$5} and $\rho=$ 1000 kg/m\textsuperscript{3}. The physical domain of interest is a 1.75$\times$1.75 mm\textsuperscript{2} square, but we consider a larger circumscribing domain to allow for wave dissipation and to avoid reflection. We pad the left edge by 2.6 mm, and the top and bottom edges by 1.75 mm. The right edge represents a line of symmetry and is not padded. We set $u_{\mathrm{R}}=$ 0.02 mm and $u_{\mathrm{I}}=$ 0 on the right edge, and $u_{\mathrm{R}} = u_{\mathrm{I}}=$ 0 on all other edges of the larger domain. We leverage a parametric prior $\prob{\X}$, which we sample from to generate realizations of the spatial distribution of the shear modulus around the ONH. The parametric prior controls the geometry of the ONH and the spatial distribution of the shear modulus. The components of the ONH that we model include the peripapillary sclera, lamina cribrosa, optic nerve, and the pia matter. We take guidance from existence literature~\cite{jonas1991morphometry, park2015lamina, vurgese2012scleral, alamouti2003retinal, yic2023ultrasonographic, hua2017intracranial} to model the geometry. \Cref{tab:onh_geom} shows the ranges of the dimensions for various components of the ONH geometry resulting from the prior. Similarly, \Cref{tab:onh_prior} shows the random variables that control the magnitude of the shear modulus field. We choose the mean and variance for these random variables based on previous studies~\cite{csahan2019evaluation, qian2021ultrasonic, zhang2020vivo}. 
	\begin{table}[t]
		\caption{Ranges of the dimensions for components of the ONH geometry resulting from the prior distribution. These ranges result from combining multiple random variables in some cases}
		\label{tab:onh_geom}
		\centering
		\begin{tabular}{lc}
			\toprule
			Random variable & Range \\
			\toprule
			Lamina cribrosa radius (width in cross-section) (mm) & $(0.55,1.35)$  \\
			Lamina cribrosa thickness (mm) & $(0.16,0.44)$ \\
			Peripapillary sclera thickness (mm) & $(0.45,1.15)$  \\
			Peripapillary sclera and retinal radius (mm) & $(8.0,16.0)$ \\
			Retinal thickness (mm) & $(0.20,0.40)$  \\
			Optic nerve radius (at bottom of cross-section) (mm) & $(1.10,2.20)$  \\
			Pia matter thickness (mm) & $(0.06,0.10)$ \\
			Rotation of the geometry (rad) & $(-\pi/12,\pi/12)$ \\
			\bottomrule
		\end{tabular}
	\end{table}
	\begin{table}[!t]
		\caption{Random variables comprising the parametric prior for the ONH. Note, $\mathcal{N}(\varpi, \varsigma^2)$ denotes a normal random variable with mean $\varpi$ and variance $\varsigma^2$}
		\label{tab:onh_prior}
		\centering
		\begin{tabular}{lc}
			\toprule
			Random variable & Distribution \\
			\toprule
			Optic nerve shear modulus (kPa) & $\mathcal{N}(9.8,3.34^2)^*$ \\
			Sclera shear modulus (kPa) & $\mathcal{N}(125,5^2)^*$ \\
			Pia matter shear modulus (kPa) & $\mathcal{N}(125,50^2)^*$ \\
			Retina shear modulus (kPa) & $\mathcal{N}(9.8,3.34^2)^*$ \\
			Lamina cribrosa shear modulus (kPa) & $\mathcal{N}(73.1,46.9^2)^*$ \\
			Background shear modulus (kPa) & $0.1$ \\
			\bottomrule
			\multicolumn{2}{c}{\small${}^{*}$The distributions are truncated to have support between $(0, 2\varsigma]$} \\
		\end{tabular}
	\end{table}

\begin{figure}[t]
		\centering
		\includegraphics[width=0.84\textwidth]{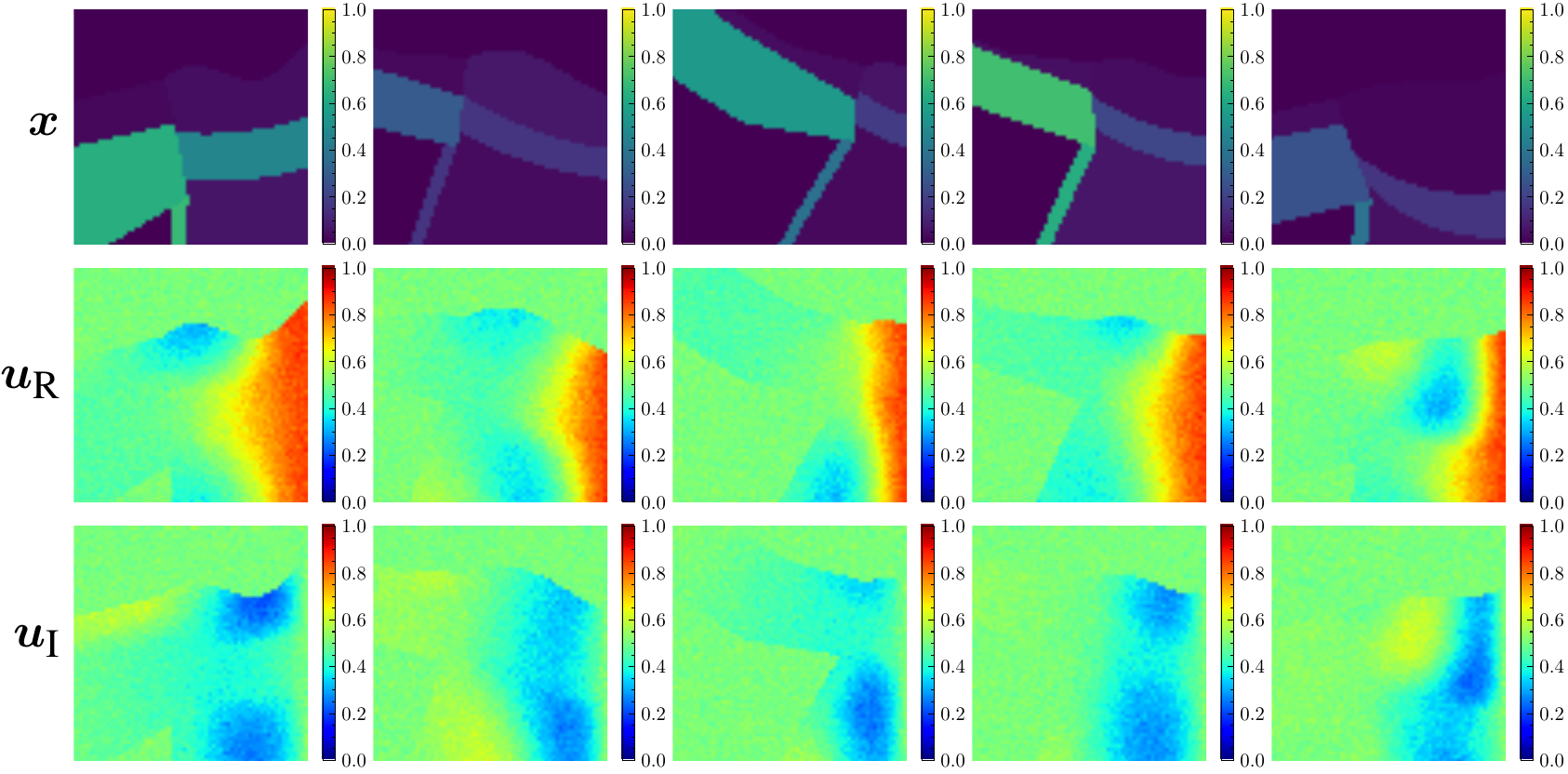}
		\caption{Five typical realizations of $\X$ and $\Y$ sampled from the joint distribution that form the training dataset for the synthetic time-harmonic elastography example. Note that $\Y$ comprises of $\uu_{\mathrm{R}}$ and $\uu_{\mathrm{I}}$ in this case. The first row shows the spatial distribution of the shear modulus field around the ONH, the second and third rows show the corresponding noisy measurements $\uu_{\mathrm{R}}$ and $\uu_{\mathrm{I}}$, respectively. All values are normalized to $[0,1]$}
		\label{fig:onh_train_samples}
	\end{figure}

	In this study, a realization of $\X$ is a shear modulus field that we discretize over a 64$\times$64 Cartesian grid. We utilize realizations of $\X$ in the finite element solver FEniCS~\cite{LoggEtal2012} to obtain noise-free predictions of the real and imaginary components of the displacement field. To this, we add homoscedastic Gaussian noise with a pre-specified standard deviation to obtain a corresponding realization of $\Y$. Therefore, the realizations of $\Y$ comprise the noisy real and imaginary components of the displacement field, which we denote using $\uu_{\mathrm{R}}$ and $\uu_{\mathrm{I}}$, respectively. Therefore, in this case, $\Nx =$ 64$\times$64 and $\Ny =$ 2$\times$64$\times$64. We build a training dataset comprising $N_{\mathrm{train}}=$ 12,000 pairs of realizations of $\X$ and $\Y$. We choose the measurement noise standard deviation to be equal to 4\% of the maximum value of $\uu_{\mathrm{R}}$ and $\uu_{\mathrm{I}}$ across the training dataset. As before, we min-max normalize the training data to $[0,1]$ for better conditioning. \Cref{fig:onh_train_samples} shows five randomly sampled training data points, where we observe significant variability in the geometry and the material property within the prior distribution. Subsequently, we train the score network using the training dataset. See \Cref{tab:hyper-parameters} for the training hyper-parameters for this problem. 
	
	\begin{figure}[!th]
		\centering
		\includegraphics[width=\textwidth]{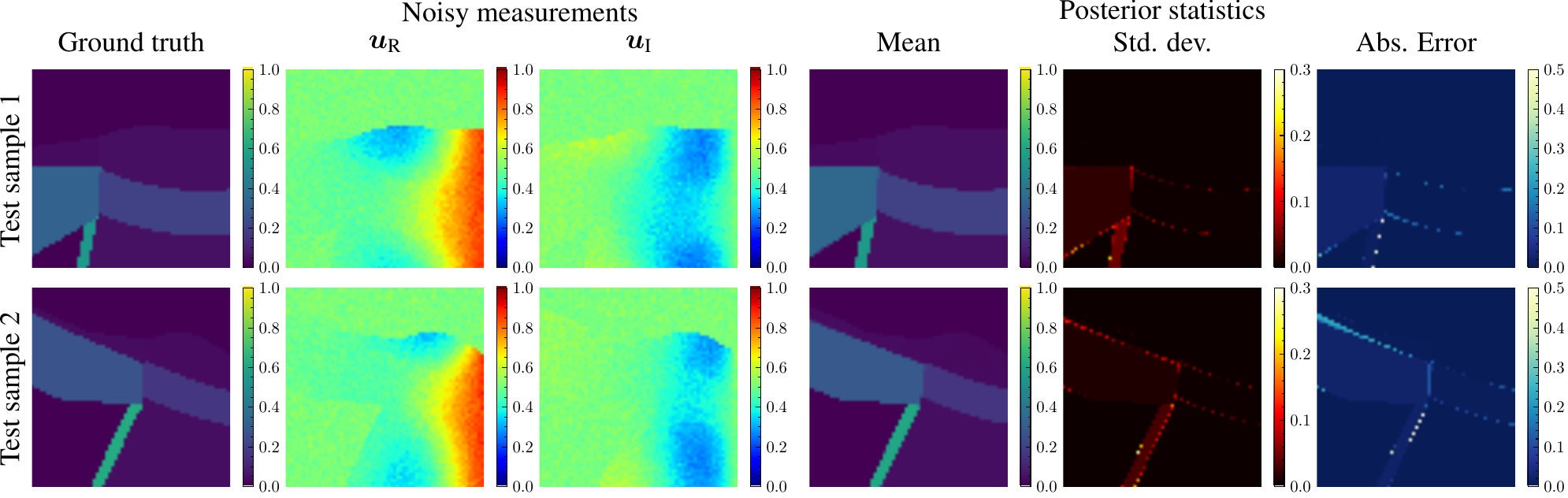}
		\caption{Posterior statistics estimated using the cSDMs for the inverse Helmholtz problem with synthetic data. Note, the ground truth distribution of the shear modulus, corresponding measurements and posterior statistics are all plotted in the normalized $[0,1]$ scale}
		\label{fig:onh_results}
	\end{figure}
	We sample the posterior distribution for two test cases that were not part of the training dataset using the trained score network and Annealed Langevin dynamics. For these cases, the first column in \Cref{fig:onh_results} shows the ground truth shear modulus distribution. The second and third columns in \Cref{fig:onh_results} show the corresponding noisy measurements. Note that the shear modulus distribution and the corresponding measurements are all plotted in the normalized $[0,1]$ scale. We sample the target posterior distribution for both cases using Annealed Langevin dynamics and generate 3,000 realizations for each case; \Cref{tab:hyper-parameters} lists the sampling hyper-parameters for this problem. We use the generated samples to estimate the posterior pixel-wise mean and standard deviation, which we plot in the fourth and fifth columns of \Cref{fig:onh_results}, respectively. The last column in \Cref{fig:onh_results} shows the absolute pixel-wise error between the posterior mean estimated using the cSDM and the corresponding ground truth. The RMSE between the posterior mean and the ground truth are 0.021 and 0.032 for the two test cases, which shows that the posterior mean provides a good reconstruction for the ground truth. We observe relatively larger uncertainty in regions of sharp change in the elastic modulus field, which also coincides with areas of larger reconstruction errors.
	
	\subsection{Applications with experimental data}\label{subsec:experimental}
	
	In this section, we apply the cSDMs to three inverse problems with experimental data. We train the score network on synthetic data but perform inference using experimental data. These examples help demonstrate the utility of the proposed approach on problems of practical relevance. 
	
	\subsubsection{Quasi-static elastography of a circular inclusion within homogeneous media}\label{subsubsec:elasto2}
	
	Similar to the example in \Cref{subsubsec:elasto}, this example also concerns an application of the proposed methodology to quasi-static elastography. We adapt this example from~\cite{ray2022efficacy}. The inverse problem at hand is that of inferring the spatial distribution of the shear modulus of a specimen from noisy full-field measurements of the vertical displacements. 
	\begin{table}[!b]
		\caption{Random variables comprising the parametric prior distribution for $\X$ in the quasi-static elastography application}
		\label{tab:elasto2_prior}
		\centering
		\begin{tabular}{lc}
			\toprule
			Random variable & Distribution \\
			\toprule
			Distance of the inclusion's center from the left edge (mm) & $\mathcal{U}(7.1,19.2)$ \\
			Distance of the inclusion's center from the bottom edge (mm) & $\mathcal{U}(7.1,27.6)$ \\
			Radius of the inclusion (mm)  & $\mathcal{U}(3.5,7.0)$ \\
			Ratio between the inclusion's and background's shear modulus (mm) & $\mathcal{U}(1,8)$ \\
			\bottomrule
		\end{tabular}
	\end{table}
	\begin{figure}[!b]
		\centering
		\includegraphics[width=0.84\textwidth]{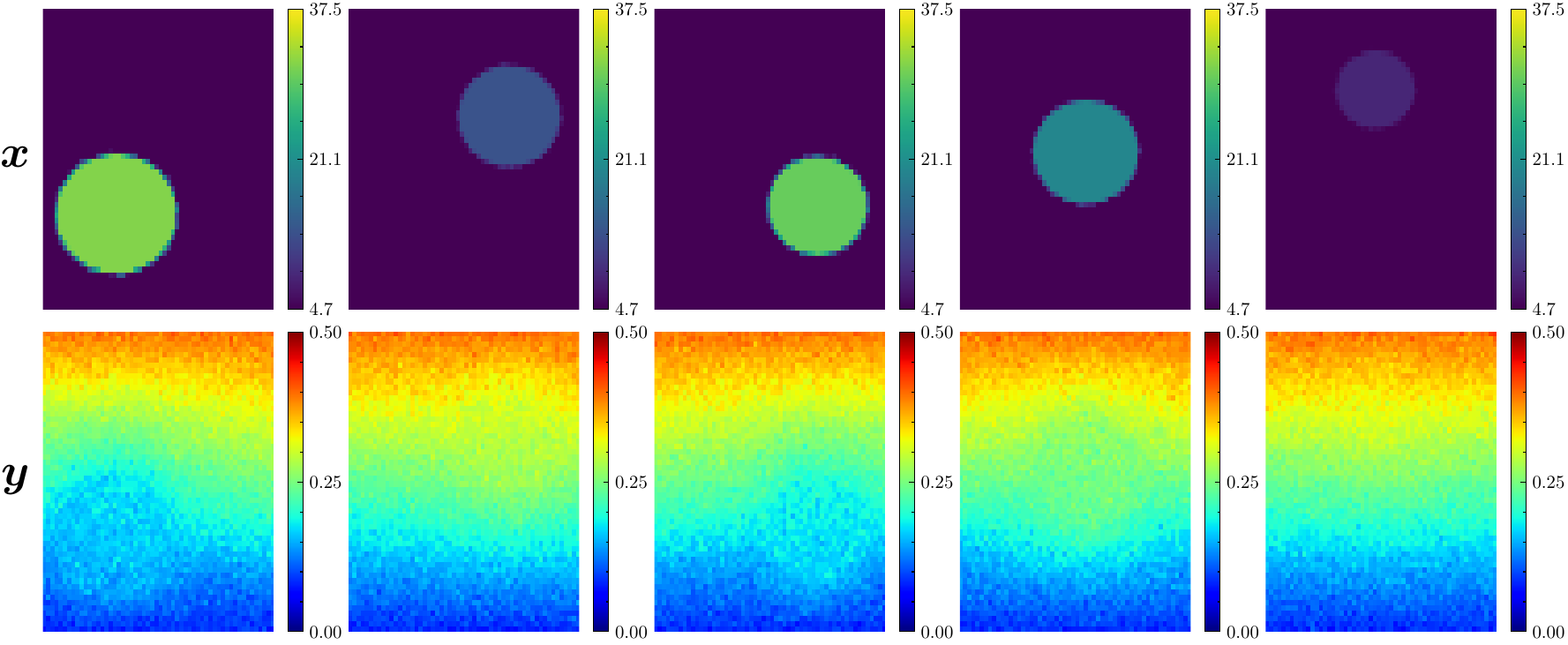}
		\caption{Five typical realizations of $\X$ and $\Y$ sampled from the joint distribution that forms the training dataset for the quasi-static elastography application. The first row shows the spatial distribution of the shear modulus field, and the second row shows the corresponding noisy measurements of the vertical component of the noisy displacement field $\uu$}
		\label{fig:elasto2_train_samples}
	\end{figure}

	For an elastic medium, the relation between the displacements and shear modulus is given by equilibrium and constitutive equations \Cref{eq:equilibrium_equation,eq:linear_constitutive_law}, respectively. We also make plane stress and incompressibility assumptions similar to previous studies~\cite{patel2021solution,ray2022efficacy}. In this example, the specimen is 34.608$\times$26.297 mm\textsuperscript{2}, with traction free boundaries along the left and right edges. Additionally, the top and bottom surfaces are traction free along the horizontal direction. To simulate compression of the specimen, the top and bottom edges are subject to 0.084 mm and 0.392 mm vertical displacements, respectively.

	In this example, a realization of $\X$ corresponds to a realization of the spatial distribution of the shear modulus of the specimen discretized over a 56$\times$56 Cartesian grid. The parametric prior for the spatial distribution of the shear modulus models a stiff circular inclusion in a uniform background of 4.7 kPa. Thus, the parametric prior consists of four random variables, which control the coordinates of the center of the inclusion, the radius of the inclusion, and the ratio of the shear modulus of the inclusion with respect to the shear modulus of the background. \Cref{tab:elasto2_prior} contains details of the parametric prior for this application~\cite{ray2022efficacy}.  We obtain a realization of $\Y$ by propagating the corresponding realization of $\X$ through a finite element model with linear triangular elements, and adding homoscedastic Gaussian noise with standard deviation equal to 0.001 mm to the predicted vertical displacement. In this manner, we build a training dataset consisting of $N_{\mathrm{train}}=8000$ pairwise realizations of $\X$ and $\Y$. \Cref{fig:elasto2_train_samples} shows five data points randomly sampled from the training dataset. Using this dataset we train the score network; see \Cref{tab:hyper-parameters} for more details regarding the various training and sampling hyper-parameters. 

\begin{figure}[t]
		\centering
		\includegraphics[width=0.85\textwidth]{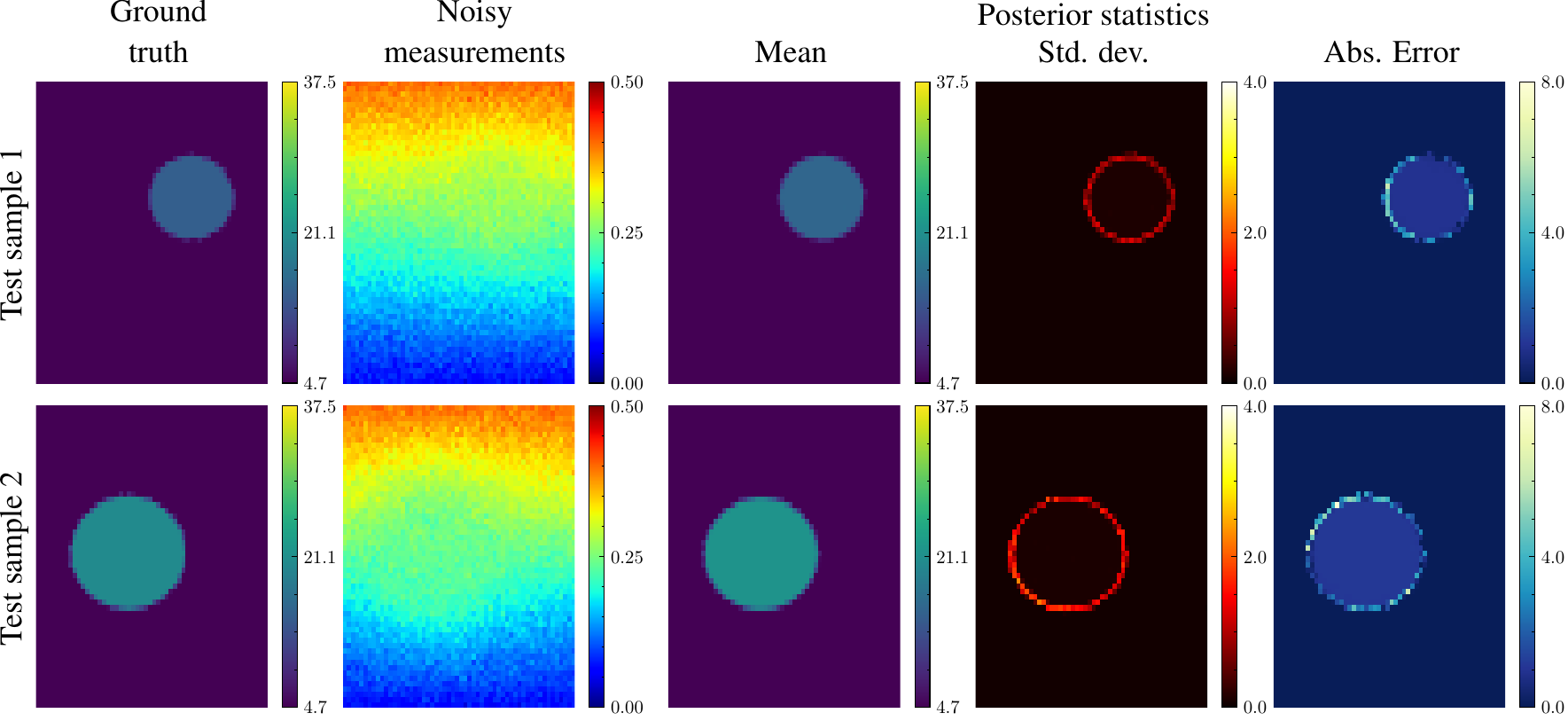}
		\caption{Posterior statistics estimated using the cSDM on test samples for the quasi-static elastography application}
		\label{fig:elasto2_test_results}
	\end{figure}

	Next, we use the trained score network and Annealed Langevin dynamics to sample the posterior distribution and infer the spatial distribution of the shear modulus for two test cases that were not part of the training dataset. \Cref{fig:elasto2_test_results} shows the results for these tests. In \Cref{fig:elasto2_test_results}, the first and second columns show the true spatial distribution for the two test cases and the corresponding full-field measurements, respectively. We use Annealed Langevin dynamics sampling to generate 800 posterior realizations for each test case and then utilize them to estimate the pixel-wise posterior mean and standard deviation.  We show the posterior statistics in the third and fourth columns of \Cref{fig:elasto2_test_results}. Finally, the last column in \Cref{fig:elasto2_test_results} shows the absolute pixel-wise error between the posterior mean estimated using the cSDM and the corresponding ground truth. The RMSE between the posterior mean and the ground truth are 0.46 and 0.63 for the two test cases. We observe relatively larger uncertainty around the edges of the stiff inclusion and comparatively less uncertainty about the stiffness of the inclusion.

	\begin{figure}[!t]
		\centering
		\includegraphics[width=0.50\textwidth]{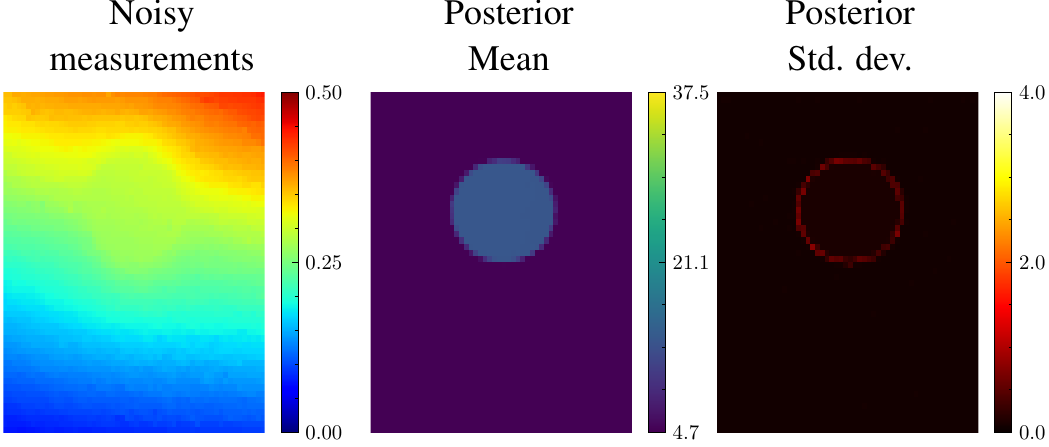}
		\caption{Posterior statistics estimated using the cSDM on experimental data for the quasi-static elastography application}
		\label{fig:elasto2_exp_results}
	\end{figure}
	Finally, we use the trained score network to infer the spatially varying shear modulus of a tissue-mimicking phantom specimen~\cite{pavan2012nonlinear}. The phantom consists of a stiff inclusion embedded within a softer substrate and was manufactured using a mixture of gelatin, agar, and oil.  The experimental measurements were obtained from ultrasound scans as the phantom was gently compressed. The left column in \Cref{fig:elasto2_exp_results} shows the measured vertical displacement. The second and third columns show the pixel-wise posterior mean and standard deviation estimate using 800 posterior samples obtained from Annealed Langevin dynamics sampling performed with the trained score network. From the posterior mean, the average stiffness and diameter of the inclusion are 12.94 kPa and 10.8 mm, respectively. These are close to the corresponding experimental measurements of 10.7 kPa and 10.0 mm, respectively. These predictions are consistent with a previous study~\cite{ray2022efficacy}.

	\subsubsection{Time-dependent elastography of a dual-layer tissue-mimicking phantom}\label{subsubsec:helmholtz2}
	This problem consists of inferring the spatial distribution of Young's modulus in a dual-layer tissue-mimicking phantom from sequential measurements of the displacement field made in an ultrasound elastography experiment~\cite{lu2021layer}. The phantom is a dual-layer cylinder of 10 mm diameter. The phantom was fabricated using 5\% gelatin concentration for the top layer and 15\% concentration for the bottom layer. The top and bottom layers were 1 mm and 3 mm thick, respectively. In the experiment, acoustic radiation force (ARF) is applied for 200~$\mu$s at sufficient depth to excite a shear wave in both layers. The ARF pushes the phantom through the central axis of the cylinder. The vertical component of the displacement field corresponding to the propagating shear wave is measured with an ultrasound transducer for the next 10~ms. The measurements are recorded at a frequency of 10~kHz and cover a two-dimensional cross-sectional plane through the center of the specimen. Based on the geometry of the specimen, the boundary conditions, and the spatial distribution of material properties and the ARF force, we assume that the problem is asymmetric about a vertical axis running through the center of the specimen. We wish to recover the Young's modulus distribution within the phantom using these measurements.
	\begin{table}[b]
		\caption{Random variables comprising the parametric prior for the time-dependent elastography application}
		\label{tab:prior_phantom}
		\centering
		\begin{tabular}{lc}
			\hline
			Parameter & Distribution \\
			\hline
			Young modulus of top phantom (kPa) & $\mathcal{U}(1,25)$ \\
			Young modulus of bottom phantom (kPa) & $\mathcal{U}(1,25)$ \\
			Height of the interface from bottom surface(mm) & $\mathcal{U}(2.0,3.6)$ \\
			\hline
		\end{tabular}
	\end{table}
	
\begin{figure}[t]
		\centering
		\includegraphics[width=\linewidth]{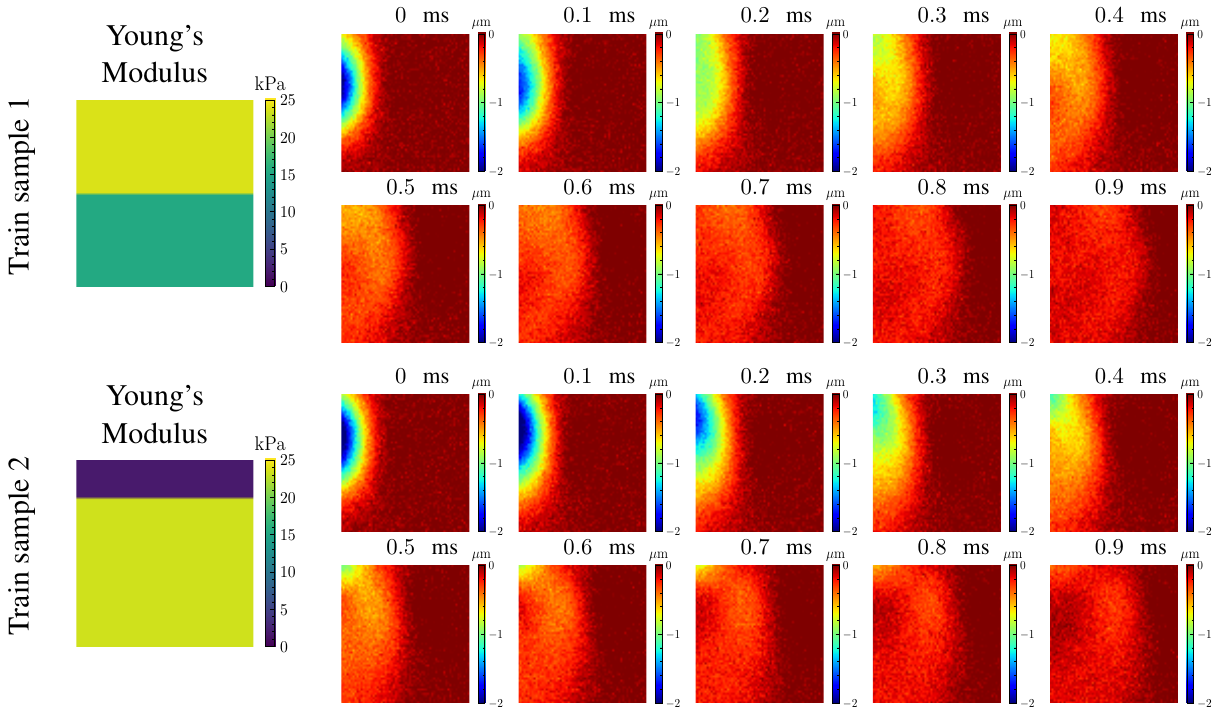}
		\caption{Two samples from the training dataset are shown. The Young's modulus distribution is shown on the left and the displacements during the $0.9$ ms after the forcing is removed are shown on the grid to the right}
		\label{fig:phantom_train_data}
	\end{figure}

	In this case, realizations of $\X$ correspond to the spatial distribution of the Young's modulus discretized on a 64$\times$64 Cartesian grid over a 3.9$\times$3.9 mm\textsuperscript{2} region. Thus, $\Nx$ =  64$\times$64 in this application. We assume that each layer in the phantom is homogeneous. Hence, the parametric prior model consists of three random variables: the Young's Modulus of the top and bottom layers and the location of the interface. See \Cref{tab:prior_phantom} for details of the parametric prior. The forward model consists of a finite element~\cite{simulia2022user} solution to the equations of elastodynamics with a linear elastic constitutive law, \Cref{eq:elastodynamic_equation,eq:linear_constitutive_law}, respectively. In this example, $\bmm$ in \Cref{eq:elastodynamic_equation} is the body force due to the ARF, and $\boldsymbol{\sigma}$ is the stress tensor. Additionally, we assume axisymmetry conditions for the specimen. Hence, we consider and plot only half of the specimen subsequently, \ie the left edge of the specimen in \Cref{fig:phantom_train_data} corresponds to the axis of rotation. We model the ARF using a body force given by
	\begin{equation}
		\bmm (s_1,s_2) = - F \exp \left\{ - \frac{s_1^2}{2l_1^2} - \frac{(s_2-\bar{s}_2)^2}{2l_2^2} \right\} \bm{e}_2,
	\end{equation}
	where $s_1$ and $s_2$ are the physical Cartesian coordinates, $\bm{e}_2$ is a unit vector in the vertical direction, $F = -$\pwr{5}{5} is a multiplicative factor used to achieve the desired magnitude of displacement, $\bar{s}_2=$ 2.6~mm is center of the ARF along the vertical direction, and $l_1 = $ 0.2~mm and $l_2 = $ 0.6~mm denote the attenuation lengths of the ARF along the horizontal and vertical directions, respectively. The ARF is applied over 200 $\mu$s to replicate the experimental setup. Once the forcing is removed, the propagation of the shear wave is modeled for 0.9~ms and ten snapshots of the vertical displacement field are recorded on a 64$\times$64 grid at a frequency of 10 kHz. This leads to $\Ny = 10 \times 64 \times 64$.
	
	We model errors in the forward model by introducing uncertainty in the modulus of the two layers and the location of the interface. This is accomplished by assigning the computed displacement field to a slightly perturbed version of the specimen for which it was calculated. The perturbed specimen is obtained by (a) scaling the modulus of each layer by a factor sampled from the uniform distribution in the interval (0.8, 1.2), and (b) translating the interface vertically by an amount sampled from a uniform distribution in the interval ($-$0.2, 0.2)~mm. These model errors are included in the forward model to account for the uncertainty in the manufacturing process. Finally, to obtain the noisy measurements, a Gaussian blurring kernel of width equal to 6 pixels is applied, and homoscedastic Gaussian noise with standard deviation equal to \pwr{5}{$-$5} mm is added. 
	In this manner, a total of 3,000 training data points are generated. \Cref{fig:phantom_train_data} shows two typical realizations of $\X$ and $\Y$ sampled from the joint distribution that are part of the training dataset. We use this dataset to train the score network and subsequently sample the posterior distribution using Annealed Langevin dynamics. 
	
	\begin{figure}[b!]
		\centering
		\includegraphics[width=\linewidth]{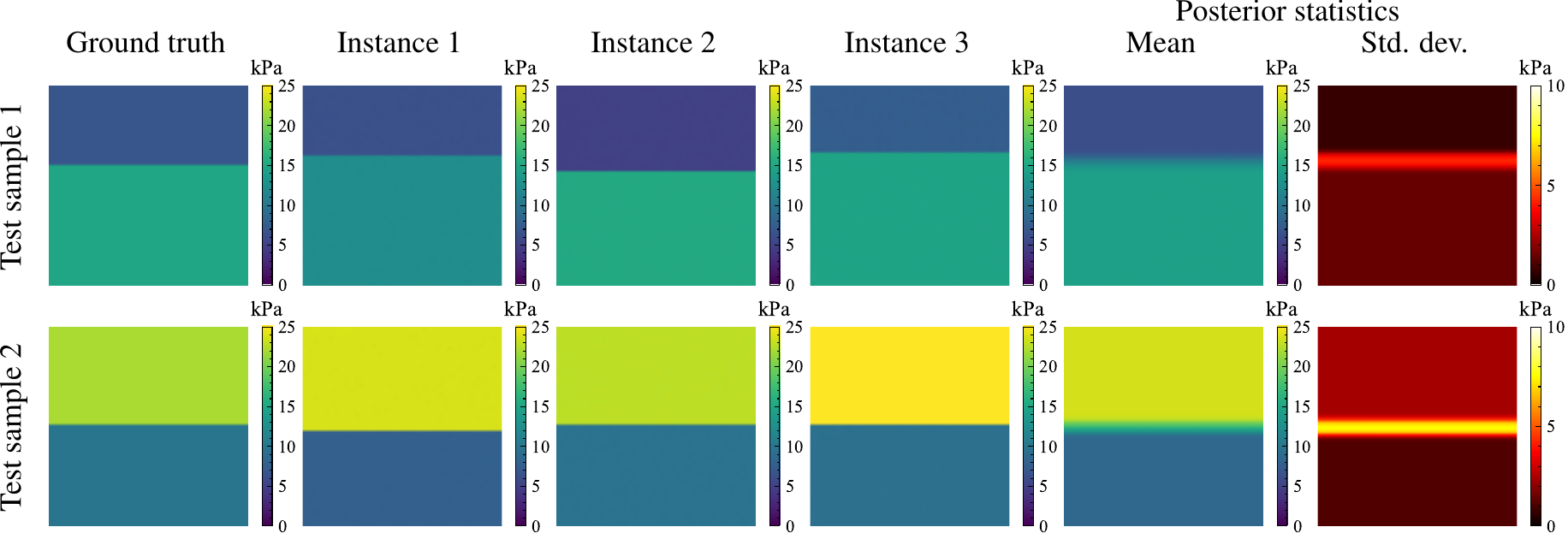}
		\caption{Posterior statistics estimated using the cSDM on test data for the time-dependent elastography application}
		\label{fig:phantom_results_synth}
	\end{figure}
	\Cref{fig:phantom_results_synth} shows the results for two test samples where we use synthetic measurements to infer the spatial distribution of the Young's moduli. The first column shows the ground truth for the Young's modulus distribution of the phantom. The second to fourth columns show three posterior realizations. The fifth and sixth columns show the pixel-wise posterior mean and standard deviation estimated using the cSDM with 500 posterior realizations. The posterior mean provides a good estimate for the ground truth both in terms of the Young's modulus for each layer and the location of the interface. \Cref{tab:phantom_test_samples} quantitatively compares the estimated posterior mean values of the Young's modulus and the location of the interface to their corresponding true values. 
	
	\begin{table}[t]
		\caption{Posterior mean estimates of the Young's modulus of the top and bottom layers of the dual-layer phantom, and the location of the interface obtained using cSDM compared to their true value for two test samples}
		\label{tab:phantom_test_samples}
		\centering
		\setlength\tabcolsep{0pt}
		\begin{tabular*}{\textwidth}{@{\extracolsep{\fill}}*{7}{c}}
			\toprule
			\multirow{2}{*}{\makecell{Test\\sample}} & \multicolumn{2}{c}{\makecell{Young's modulus of\\top layer (kPa)}} & \multicolumn{2}{c}{\makecell{Young's modulus of\\bottom layer (kPa)}} & \multicolumn{2}{c}{\makecell{Location of\\interface (mm)}} \\
			\cline{2-3} \cline{4-5} \cline{6-7}
			& \makecell{True\\value} & \makecell{Posterior\\mean} & \makecell{True\\value} & \makecell{Posterior\\mean} & \makecell{True\\value} & \makecell{Posterior\\mean}  \\
			\hline
			1 & 14.74& 14.20& 6.66& 6.01& 2.48 & 2.57 \\
			2 & 21.81& 23.42& 9.70& 8.48&  2.11 & 2.05 \\
			\hline
		\end{tabular*}
	\end{table}
	\begin{figure}[H]
		\centering
		\includegraphics[width=\linewidth]{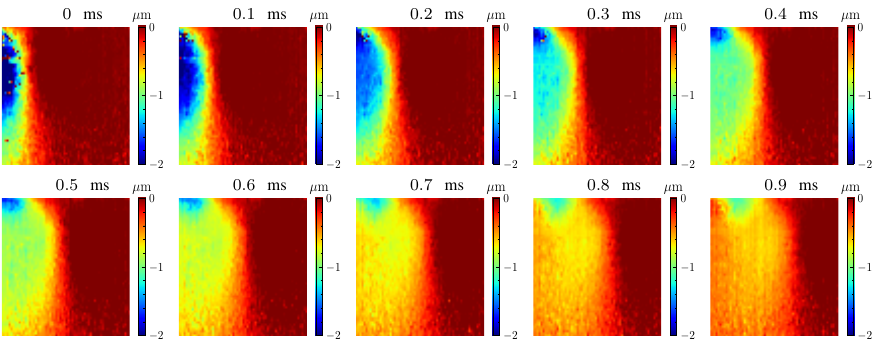}
		\caption{Ultrasound measurements of the displacement in $\mu$m of a dual layer phantom during an interval of $0.9$ ms. Data collected by~\citet{lu2021layer}}
		\label{fig:phantom_meas}
	\end{figure}
	\begin{figure}[H]
		\centering
		\includegraphics[width=\linewidth]{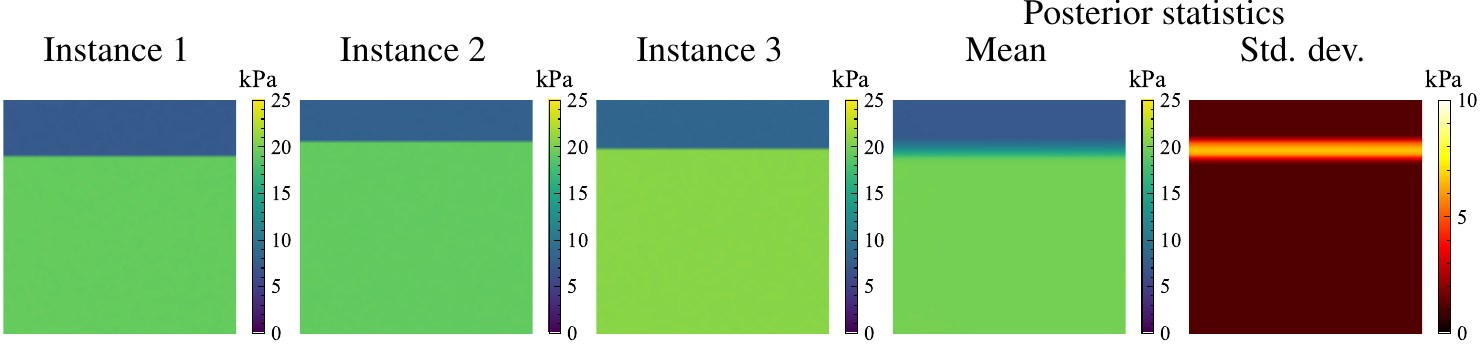}
		\caption{Posterior realizations (first three columns) and statistics estimated using the cSDM on experimental data for the ultrasound elastography application}
		\label{fig:phantom_results_meas}
	\end{figure}
	
	\Cref{fig:phantom_results_meas} shows the results of the inference for the experimental data collected by \citet{lu2021layer} and shown in \Cref{fig:phantom_meas}. This figure shows three typical samples from the posterior distribution and the pixel-wise mean and standard deviation computed using  500 realizations from the posterior distribution. The posterior mean estimate for the Young's modulus of the top and bottom layers is 6.98  kPa and 19.73 kPa, respectively. Uniaxial quasi-static mechanical tests performed on homogenous phantoms with the same gelatin concentrations as the two layers yielded values of 2.1 $\pm$ 0.2 kPa and 20.3 $\pm$ 0.4 kPa. Thus, the cSDM estimates the modulus of the lower layer with good accuracy. However, there is a significant error in the estimate for the modulus of the top layer. We observed a similar overestimation in the Young's modulus of a thin top layer on a test example similar to the phantom but not part of the training dataset. There are two plausible sources for this error. The overestimation of the Young's modulus of a thin top layer may be because its contribution to the measured displacement is unlikely to be significant. Hence, the probabilistic model for modeling errors, which includes discrepancies in the forward physics and measurement noise, may require revision to account for such atypical cases. Alternatively, the erroneous prediction may be due to a lack of similar examples in the training dataset and the score network not generalizing well to such cases. The phantom was fabricated such that the interface would be at a height of 3.0 mm. The estimate for the mean interface height obtained using the cSDM is 3.21~mm with an average standard deviation of 0.10 mm; the true location of the interface is close to two standard deviations of the estimated posterior mean. 
	\subsubsection{Optical coherence elastography of tumor spheroids}\label{subsubsec:TS}
	
	\begin{figure}[b]
		\centering
		\includegraphics[height=2in]{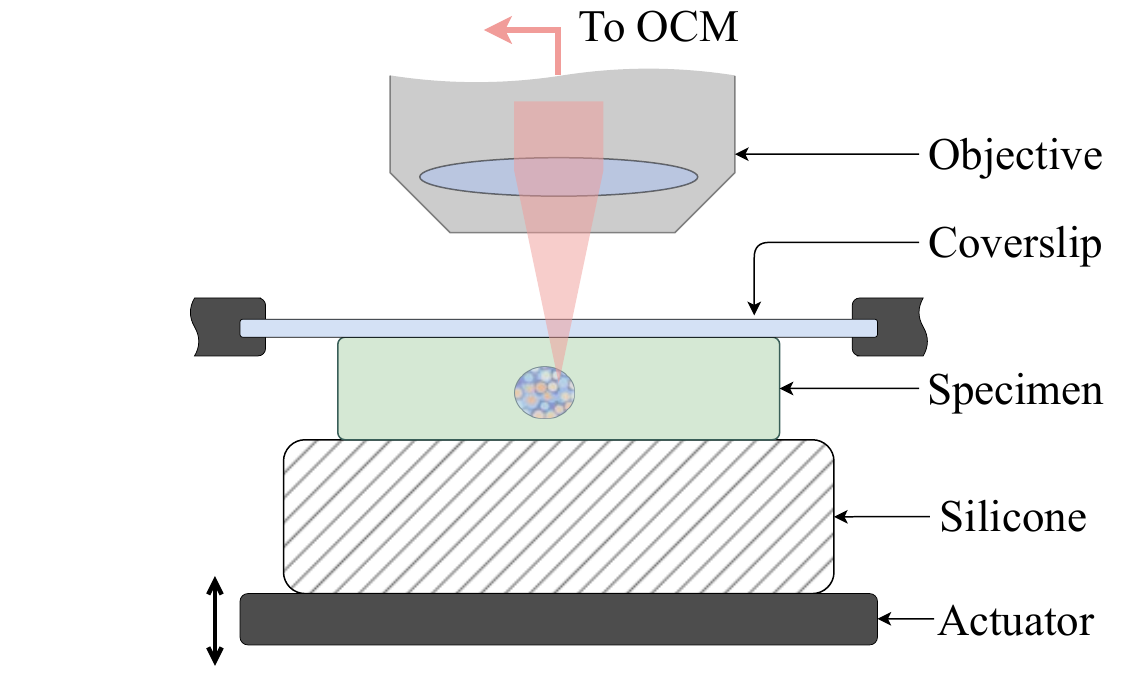}
		\caption{Schematic showing the experimental setup in the tumor spheroid application}
		\label{fig:ts_experimental_setup}
	\end{figure}
	The final application concerns OCE--an optical coherence tomography (OCT) method that enables mechanical contrast-based imaging using the phase data of backscattered light from a specimen undergoing compression~\cite{kennedy2014review,kennedy2014optical}. More specifically, this application considers mechano-microscopy~\cite{mowla2022subcellular,mowla2024multimodal} of tumor spheroids. Mechano-microscopy is a high resolution variant of phase-sensitive compression OCE, which utilizes a high-resolution variant of OCT known as optical coherence microscopy (OCM). Tumor spheroids refer to a collection of multiple tumor cells. We provide a short background on OCE in \Cref{appsubsec:TS_background} and refer interested readers to \cite{olofsson2018acoustic} for typical three-dimensional immuno-fluorescence images of tumor spheroids taken at different time points. In this application, we use a cSDM to infer the spatial distribution of the Young's modulus $\X$ of tumor spheroids from noisy phase difference measurements $\Y$, which encode information regarding the specimen's displacement along the optical axis. \Cref{fig:ts_experimental_setup} provides a schematic of the experimental setup. We relegate additional details regarding the physical experiment (\Cref{appsubsec:TS_exp_details}) and specimen preparation process (\Cref{appsubsec:TS_specimen})  to \ref{app:TS_details}. The data for this experiment were obtained from measurements and datasets originally compiled by~\citet{foo2024tumor}.

	\begin{figure}[t]
		\centering
		\includegraphics[width=\textwidth]{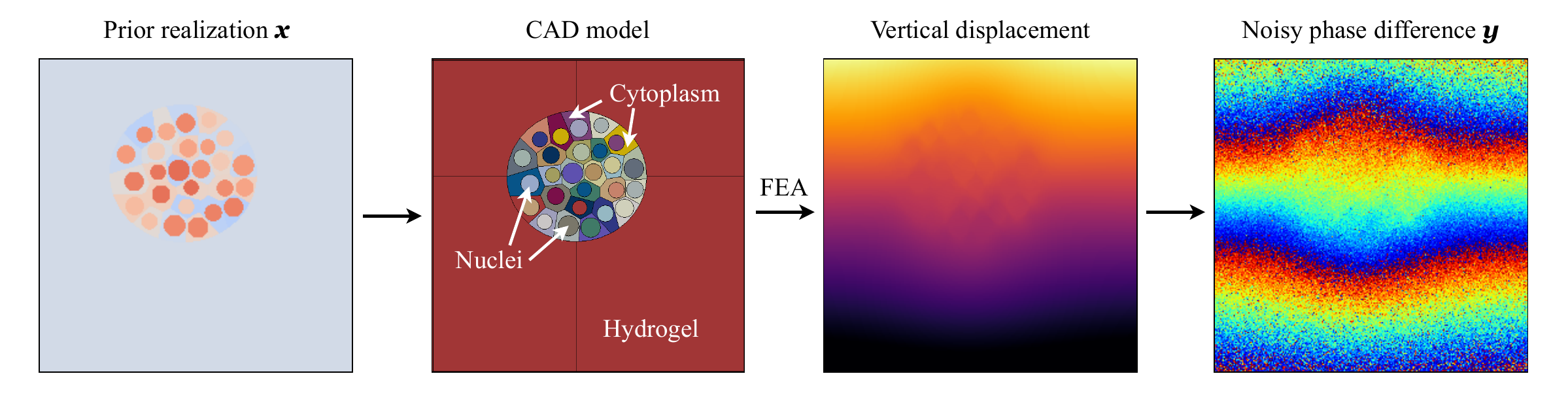}
		\caption{Workflow for generating the training data in the tumor spheroid application}
		\label{fig:ts_schematic}
	\end{figure}
	The synthetic samples, which we generate from the parametric prior distribution, represent a spheroid within a hydrogel on a 256$\times$256~$\mu$m\textsuperscript{2} domain (first column in \Cref{fig:ts_schematic}).  We describe the parametric prior for this application in \Cref{appsubsec:TS_prior}. The parametric prior is highly complex, involving a custom nuclei placement algorithm and Voronoi tessellations~\cite{imai1985voronoi}. We discretize the Young's modulus field on a 256$\times$256 Cartesian grid over the specimen.  We record noisy measurements of the wrapped phase difference field over the same grid. \Cref{fig:ts_schematic} shows the steps for generating realizations of $\Y$ from realizations of $\X$. We provide details of the forward model in \Cref{appsubsec:TS_fwdmodel}, which involves developing a CAD model for the specimen, finite element analysis (FEA) using ABAQUS~\cite{simulia2022user}, and a non-additive non-Gaussian measurement noise model (\Cref{appeq:TS_noisemodel}). To be precise, every realization of $\X$ is a Young's modulus field, and the corresponding realization of $Y$ is the wrapped and noisy phase difference field. Also, $\Nx = \Ny = $ 256$\times$256 for this application.
	
\begin{figure}[t]
		\centering
		\includegraphics[width=\textwidth]{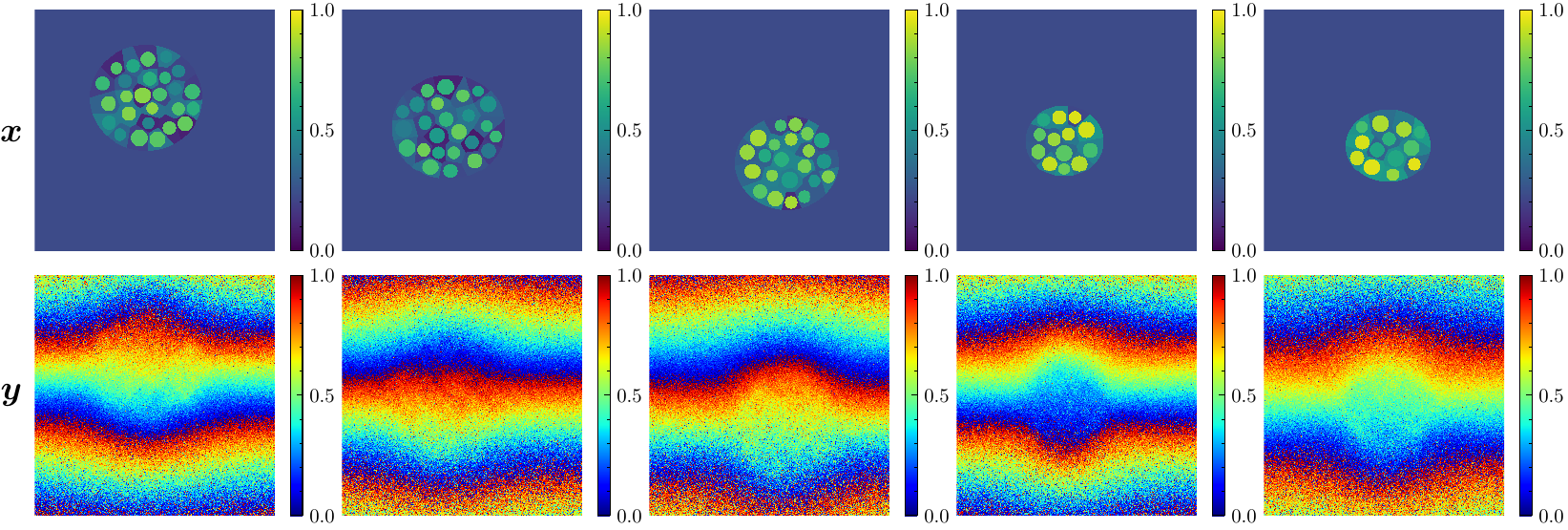}
		\caption{Five typical realizations of $\bm{X}$ and $\bm{Y}$ sampled from the joint distribution that form the training dataset for the tumor spheroid experiment. In the first row, we have the log-normalized Young's module fields, and in the second row, the corresponding noisy measurements are displayed. All values are normalized to $[0,1]$ scale}
		\label{fig:ts_sample}
	\end{figure}

	We initially synthesize a synthetic dataset comprising 5000 pairs of $\bm{X}$ and $\bm{Y}$ realizations. We augment this dataset  sixfold by shifting the 256$\times$256~$\mu\text{m}^2$ image window 20~$\mu$m in each horizontal and vertical direction (left, right, up, and down), as well as applying a horizontal flip to each image. We divide the resulting dataset into a training set of 24000 images and a test set of 6000 images. As a preprocessing step, we rescale the elasticity images to enhance contrast and facilitate better training convergence. The rescaling involves the following transformation
	\begin{equation}
		E^\prime = \log_{10} \left( \frac{E}{E_{\text{hydrogel}}}\right),
	\end{equation}
	where $E^\prime$ is the rescaled Young's modulus. This transformation means that it will only be possible to recover the Young's modulus field up to the normalization constant $E_{\text{hydrogel}}$, which is the Young's modulus of the hydrogel. \Cref{fig:ts_sample} illustrates five randomly selected samples from the dataset. We utilize the training data to train the score network. After training the score network, we employ Annealed Langevin dynamics to sample elasticity images from the posterior. See \Cref{tab:hyper-parameters} for additional information regarding the training and sampling hyperparameters.  
	
	\begin{figure}[!t]
		\centering
		\includegraphics[width=0.83\textwidth]{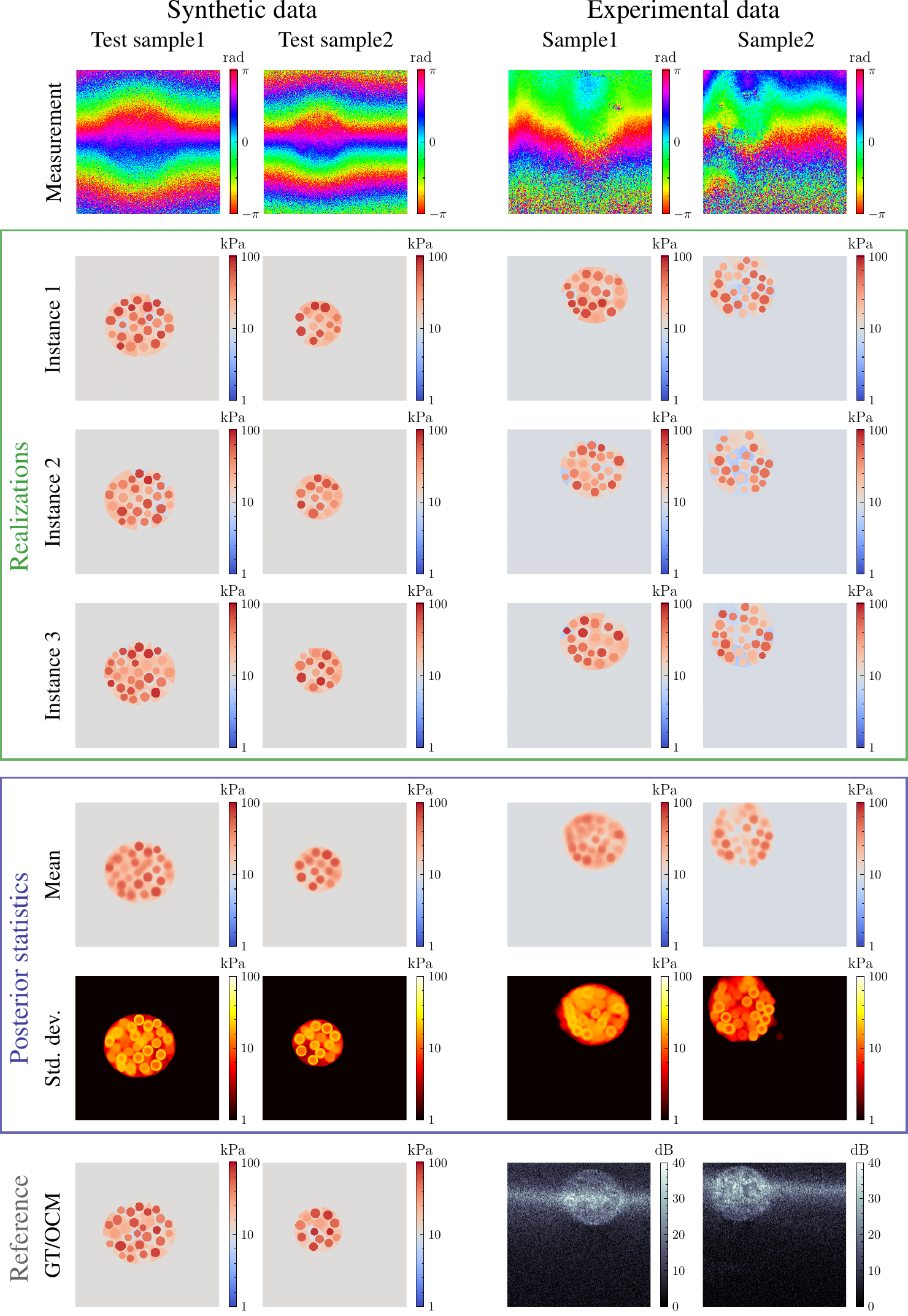}
		\caption{Posterior statistics estimated for selected samples using the cSDM for the tumor spheroid problem. The field of view in each image is 256$\times$256~$\mu\text{m}^2$}
		\label{fig:ts_results}
	\end{figure}
	\Cref{fig:ts_results} shows realizations from the posterior, and its statistics, estimated using the cSDM on both synthetic and experimental data. The first two columns correspond to samples from the test dataset previously unseen by the model, while columns three and four depict predictions on experimental measurements. The first row shows the measurements used for sampling using the trained score network. Rows two to four show three realizations sampled by the model for the corresponding measurements. Rows five and six show the pixel-wise posterior mean and standard deviation, respectively, estimated using 200 realizations. For the synthetic samples, we compare the estimated posterior means for the Young's modulus distribution with the corresponding ground truth. In the case of experimental data, comparisons were made against OCM images that were co-registered during the acquisition of the phase difference images. Note that all images are presented at their true scales. Recall that the recovery of the Young's modulus is possible up to a multiplicative constant: the hydrogel modulus $E_{\text{hydrogel}}$ in this case. We assume $E_{\text{hydrogel}} = $ 10~kPa for the test samples. For the experimental samples, $E_{\text{hydrogel}}$ was estimated from mechano-microscopy measurements~\cite{foo2024tumor}.
	
	From the results for the synthetic data, we note that each realization from the posterior distribution correctly captures the volume and the location of the spheroid. Further, there are significant differences between each realization in the modulus distribution predicted within each spheroid. We can attribute these differences to the ill-posed nature of the inverse problem \cite{barbone2004elastic,ferreira2012uniqueness}, which stipulates that there are multiple modulus distributions that are consistent with a given noisy measured displacement field. The differences in the modulus field within the spheroid lead to a fuzzy mean modulus field. Remarkably, when we compare this field to the ground truth, we observe that the location of several nuclei, which correspond to regions with elevated stiffness, is captured accurately. The uncertainty in the modulus field within the spheroid directly translates to large values of standard deviation within the spheroid. We also observe that regions of the largest standard deviation (uncertainty) are associated with the boundaries of the nuclei.
	
	From the results for the experimental data, we observe that the measured phase field is more heterogeneous and noisy than its synthetic counterpart. Thus, we anticipate more variability and uncertainty in the predictions, which is precisely what we observe when examining the three realizations from the posterior distribution. This translates to a fuzzy image for the mean stiffness field and large standard deviation values (when compared with synthetic data) within the spheroid. The underlying ground truth image is unavailable in the experimental case. Therefore, we cannot quantitatively evaluate the accuracy of the modulus field. However, from the OCM images (last row of \Cref{fig:ts_results}), we can infer the location and the size of the spheroid, and when we compare this with the mean modulus image we conclude that the location and the size of the spheroid have been captured accurately.
	
	We conclude this section by noting that although the initial tumor spheroid results are encouraging, they point to several extensions that are currently underway. These include improving the forward model for measurement noise so that the synthetic measurements resemble their experimental counterparts more closely and performing inference on tissue-mimicking phantoms of tumor spheroids for a quantitative comparison between the inferred results and the ground truth.
	
	\section{Conclusions, outlook, and extensions}\label{sec:conclusion}
	
	Large-scale Bayesian inference using black-box models is challenging. In this work, we explore the use of cSDMs for large-scale Bayesian inference. Fundamental to the score-based diffusion models is the score network that can be trained using realizations from the joint distribution between the inferred quantity and the measurements. Thus, we only need to simulate from black-box models to create the training dataset. The trained score network can be used to sample the target posterior distribution using Annealed Langevin dynamics. The score network learns the score function of the target posterior  `\textit{averaged}' over the joint distribution. Therefore, there is no need to re-train the score function when a new measurement is made. This makes cSDMs particularly suitable for high-throughput applications. 
	
	We apply cSDMs to solve several inverse problems in mechanics. We validate the proposed approach on two synthetic examples; for these problems the quality of the inference is similar to MCS.  Further, we demonstrate the utility of the proposed approach on three applications where experimental measurements are available. Overall, we show that cSDMs can tackle high-dimensional inverse problems involving complex prior distributions, different types of forward physics simulated using black-box models, complex measurement noise models and modeling errors, and various measurement modalities.
	
	Although the results in this manuscript are promising, we identify several interesting future research directions that will help with the adoption of this technology as an inference toolbox. First, the primary computational bottleneck in applying cSDMs, and any conditional generative algorithm for that matter, will be the cost of curating the training dataset, which requires multiple simulations of the black-box forward model. Thus, there is a need for the development of data-efficient conditional diffusion models. Moreover, training using multi-fidelity datasets promises to be another interesting research direction. Further, we consider full-field measurements in this work. Extensions to sparse measurements using masking-based training strategies~\cite{he2022masked} is another extension that will increase the utility of the proposed method. We also assume that models for the measurement noise and modeling errors are known because we need to sample from them when generating realizations of the observation $\Y$ from corresponding realizations of $\X$. However, complete knowledge of the measurement noise and/or modeling errors may not available  \latin{a priori}. In such cases, the inverse problem may require hyperpriors over parameters of the noise or error models, or require careful considerations of epistemic uncertainty surrounding them. Extending the capabilities of the current framework to handling hyperpriors or epistemic uncertainty is yet another interesting future research direction.
	
	Finally, applying cSDMs to physics-based inverse problems that arise in different science and engineering domains will be interesting, as it will reveal insights into the behavior of conditional generative algorithms of this type. Quantifying the reliability of the inference results obtained using conditional generative models is also important. Evaluating the trained conditional generative model using posterior predictive checks and cross validation is necessary~\cite{gelman2020bayesian}. However, the required auxiliary datasets to perform posterior predictive checks or the computational budget necessary for cross validation may be unavailable. Alternative approaches to assessing estimates of aleatoric uncertainties--primary sources of uncertainty in inverse problems of the type we consider--similar to metrics to evaluate epistemic uncertainty that stems from neural network models~\cite{xiong2024uncertainty} are also needed.
	
	\section{Acknowledgments}
	The authors acknowledge support from ARO, United States, USA grant W911NF2010050, and from the National Institutes of Health grant 1R01EY032229-01. BFK acknowledges funding from the Australian Research Council, the NAWA Chair programme of the Polish National Agency for Academic Exchange and from the National Science Centre, Poland. The authors acknowledge the following colleagues for their contribution to the results presented in \Cref{subsubsec:TS}: Alireza Mowla, Matt Hepburn, Jiayue Li, Danielle Vahala, Sebastian Amos, Liisa Hirvonen, Samuel Maher, and Yu Suk Choi. The authors also acknowledge the Center for Advanced Research Computing (CARC, \href{https://carc.usc.edu}{carc.usc.edu}) at the University of Southern California for providing computing resources that have contributed to the research results reported within this publication.
	
	\bibliographystyle{elsarticle-num-names} 
	\bibliography{references}
	
	\appendix
	\setcounter{section}{0}
	\renewcommand{\appendixname}{Appendix}
	\renewcommand{\thesection}{Appendix~\Alph{section}}
	\renewcommand{\thesubsection}{\Alph{section}\arabic{subsection}}
	\renewcommand{\thefigure}{\Alph{section}\arabic{figure}}
	\setcounter{figure}{0}
	\renewcommand{\thetable}{\Alph{section}\arabic{table}}
	\setcounter{table}{0}
	
	\section{Derivation of \Cref{eq:DSM-2} from \Cref{eq:dsm1}}\label{app:derivation}
	
	Using \Cref{eq:convolution},
	\begin{equation}\label{appeq:convolution}
		\prob{\tilde{\X}\vert\Y}(\tilde{\x} \vert \y) = \int \probb{\tilde{\X}\vert\X}{\tilde{\x}\vert\x} \; \prob{\X\vert\Y}(\x \vert \y) \; \mathrm{d}\x,
	\end{equation}
	we wish to show the denoising score matching objective \Cref{eq:dsm1}
	\begin{equation}\label{appeq:dsm1}
		\ell_{\mathrm{DSM}}(\y; \thetaa) =  \frac{1}{2}\Exp_{\tilde{\X} \sim \prob{\tilde{\X}\vert\Y}} \Big[ \lVert  \sm(\tilde{\x}, \y; \thetaa) - \nabla_{\tilde{\x}} \log  \prob{\tilde{\X}\vert\Y}\!(\tilde{\x}\vert\y)\rVert_2^2 \Big],
	\end{equation}
	is equivalent to \Cref{eq:DSM-2}, \ie
	\begin{equation}\label{appeq:DSM-2}
		\ell_{\mathrm{DSM}}(\y; \thetaa) = \frac{1}{2} \Exp_{\X \sim \prob{\X\vert\Y}} \Bigg[ \Exp_{\tilde{\X} \sim \prob{\tilde{\X}\vert\X}} \Big[ \lVert  \sm(\tilde{\x}, \y; \thetaa) - \nabla_{\tilde{\x}} \log  \probb{\tilde{\X}\vert\X}{\tilde{\x}\vert\x}\rVert_2^2 \Big] \Bigg] + \mathcal{C},
	\end{equation} 
	where $\mathcal{C}$ is an additive constant that does not depend on $\thetaa$. To show this, we start with \Cref{appeq:dsm1} and expand the integrand
	\begin{equation}\label{appeq:DSM-2-part1}
		\begin{split}
			\ell_{\mathrm{DSM}}(\y; \thetaa) &=  \frac{1}{2}\Exp_{\tilde{\X} \sim \prob{\tilde{\X}\vert\Y}} \Big[ \lVert  \sm(\tilde{\x}, \y; \thetaa) - \nabla_{\tilde{\x}} \log  \prob{\tilde{\X}\vert\Y}\!(\tilde{\x}\vert\y)\rVert_2^2 \Big] \\
			&= \frac{1}{2}\Exp_{\tilde{\X} \sim \prob{\tilde{\X}\vert\Y}} \Big[ \Big< \sm(\tilde{\x}, \y; \thetaa), \sm(\tilde{\x}, \y; \thetaa)\Big> - 2 \Big< \sm(\tilde{\x}, \y; \thetaa), \nabla_{\tilde{\x}} \log  \prob{\tilde{\X}\vert\Y}\!(\tilde{\x}\vert\y) \Big> \\
			&\hphantom{= \frac{1}{2}\Exp_{\tilde{\X} \sim \prob{\tilde{\X}\vert\Y}} \Big[ \Big< \sm(\tilde{\x}, \y; \thetaa), \sm(\tilde{\x}, \y; \thetaa)\Big>} + \Big< \nabla_{\tilde{\x}} \log  \prob{\tilde{\X}\vert\Y}\!(\tilde{\x}\vert\y), \nabla_{\tilde{\x}} \log  \prob{\tilde{\X}\vert\Y}\!(\tilde{\x}\vert\y) \Big> \Big],
		\end{split}
	\end{equation}
	where $<\cdot, \cdot>$ denotes the Euclidean inner product in $\Re^\Nx$, \ie $<\am, \am> \,= \am\transpose\am$ . Note that the last term \Cref{appeq:DSM-2-part1} does not depend on $\thetaa$, so, herein, we will denote the last term using $\mathcal{C}_1^\prime$. Now, we expand the expectations in \Cref{appeq:DSM-2-part1} and utilize \Cref{appeq:convolution} to rewrite \Cref{appeq:DSM-2-part1} as follows:
	\begin{equation}\label{appeq:DSM-2-part2}
		\begin{split}
			\ell_{\mathrm{DSM}}(\y; \thetaa) &=  \frac{1}{2} \Bigg\{ \left[ \int \Big< \sm(\tilde{\x}, \y; \thetaa), \sm(\tilde{\x}, \y; \thetaa)\Big> \; \probb{\tilde{\X}\vert\Y}{\tilde{\x}\vert\y} \mathrm{d}\tilde{\x} \right] \\
			&\hphantom{= \sm(\tilde{\x}, \y; \thetaa) } - 2 \left[ \int \Big< \sm(\tilde{\x}, \y; \thetaa), \nabla_{\tilde{\x}} \log  \prob{\tilde{\X}\vert\Y}\!(\tilde{\x}\vert\y) \Big> \; \probb{\tilde{\X}\vert\Y}{\tilde{\x}\vert\y} \mathrm{d}\tilde{\x} \right] \Bigg\} + \mathcal{C}_1,
		\end{split}
	\end{equation}
	where $\mathcal{C}_1 = \frac{1}{2} \Exp_{\tilde{\X} \sim \prob{\tilde{\X}\vert\Y}} \{ \mathcal{C}_1^\prime \}$. Utilizing
	\begin{equation}
		\nabla_{\tilde{\x}} \log  \prob{\tilde{\X}\vert\Y}\!(\tilde{\x}\vert\y) = \frac{1}{\prob{\tilde{\X}\vert\Y}\!(\tilde{\x}\vert\y)} \nabla_{\tilde{\x}} \prob{\tilde{\X}\vert\Y}\!(\tilde{\x}\vert\y),
	\end{equation}
	\begin{equation}
		\nabla_{\tilde{\x}} \log  \prob{\tilde{\X}\vert\X}\!(\tilde{\x}\vert\x) = \frac{1}{\prob{\tilde{\X}\vert\X}\!(\tilde{\x}\vert\x)} \nabla_{\tilde{\x}} \prob{\tilde{\X}\vert\X}\!(\tilde{\x}\vert\x),
	\end{equation}
	and \Cref{appeq:convolution}, we can simplify the second term in \Cref{appeq:DSM-2-part2} as follows:
	\begin{equation}\label{appeq:DSM-2-part3}
		\begin{split}
			&\int \Big< \sm(\tilde{\x}, \y; \thetaa), \nabla_{\tilde{\x}} \log  \prob{\tilde{\X}\vert\Y}\!(\tilde{\x}\vert\y) \Big> \probb{\tilde{\X}\vert\Y}{\tilde{\x}\vert\y} \mathrm{d}\tilde{\x} = \int \Big< \sm(\tilde{\x}, \y; \thetaa), \nabla_{\tilde{\x}}  \prob{\tilde{\X}\vert\Y}\!(\tilde{\x}\vert\y) \Big> \;\mathrm{d}\tilde{\x} \\
			&\hphantom{  \log \log \log   }= \int \Big< \sm(\tilde{\x}, \y; \thetaa), \nabla_{\tilde{\x}}  \underbrace{\int \probb{\tilde{\X}\vert\X}{\tilde{\x}\vert\x} \; \prob{\X\vert\Y}(\x \vert \y) \; \mathrm{d}\x}_{\prob{\tilde{\X}\vert\Y}\!(\tilde{\x}\vert\y)} \Big> \;\mathrm{d}\tilde{\x} \\
			&\hphantom{ \log \log \log   }= \int \int \Big< \sm(\tilde{\x}, \y; \thetaa), \nabla_{\tilde{\x}}  \log \probb{\tilde{\X}\vert\X}{\tilde{\x}\vert\x} \Big> \; \probb{\tilde{\X}\vert\X}{\tilde{\x}\vert\x} \prob{\X\vert\Y}(\x \vert \y) \;\mathrm{d}\tilde{\x}\,\mathrm{d}\x, 
		\end{split}
	\end{equation}
	where we make use of the homogeneity of the inner product operator, \ie $<\am; \alpha \am> = \alpha <\am ; \am> \; \forall \; \am \in \Re^{\Nx}$ and $\alpha \in \Re$, when simplifying the initial expression and then again in the last step. On substituting \Cref{appeq:DSM-2-part3} in \Cref{appeq:DSM-2-part2} and expanding the first term, we obtain
	\begin{equation}\label{appeq:DSM-2-part4}
		\begin{split}
			\ell_{\mathrm{DSM}}(\y; \thetaa) &=  \frac{1}{2} \Bigg\{ \left[ \int\int \Big< \sm(\tilde{\x}, \y; \thetaa), \sm(\tilde{\x}, \y; \thetaa)\Big>  \probb{\tilde{\X}\vert\X}{\tilde{\x}\vert\x} \prob{\X\vert\Y}(\x \vert \y) \;\mathrm{d}\tilde{\x}\,\mathrm{d}\x \right] \\
			&\hphantom{= } - 2 \left[ \int \int \Big< \sm(\tilde{\x}, \y; \thetaa), \nabla_{\tilde{\x}}  \log \probb{\tilde{\X}\vert\X}{\tilde{\x}\vert\x} \Big> \probb{\tilde{\X}\vert\X}{\tilde{\x}\vert\x} \prob{\X\vert\Y}(\x \vert \y) \;\mathrm{d}\tilde{\x}\,\mathrm{d}\x \right] \Bigg\} + \mathcal{C}_1.
		\end{split}
	\end{equation}
	To the right hand side in \Cref{appeq:DSM-2-part4}, we add and subtract another term that does not depend on $\thetaa$
	\begin{equation}
		\mathcal{C}_2 = \frac{1}{2} \int \int \Big< \nabla_{\tilde{\x}} \log  \prob{\tilde{\X}\vert\X}\!(\tilde{\x}\vert\x), \nabla_{\tilde{\x}} \log  \prob{\tilde{\X}\vert\X}\!(\tilde{\x}\vert\x) \Big> \; \probb{\tilde{\X}\vert\X}{\tilde{\x}\vert\x} \prob{\X\vert\Y}(\x \vert \y) \;\mathrm{d}\tilde{\x}\,\mathrm{d}\x.
	\end{equation}
	Now,
	\begin{equation}\label{appeq:DSM-2-part5}
		\begin{split}
			\ell_{\mathrm{DSM}}(\y; \thetaa) &=  \frac{1}{2} \Bigg\{ \left[ \int\int \Big< \sm(\tilde{\x}, \y; \thetaa), \sm(\tilde{\x}, \y; \thetaa)\Big> \;  \probb{\tilde{\X}\vert\X}{\tilde{\x}\vert\x} \prob{\X\vert\Y}(\x \vert \y) \;\mathrm{d}\tilde{\x}\,\mathrm{d}\x \right] \\
			&\hphantom{= } - 2 \left[ \int \int \Big< \sm(\tilde{\x}, \y; \thetaa), \nabla_{\tilde{\x}}  \log \probb{\tilde{\X}\vert\X}{\tilde{\x}\vert\x} \Big> \; \probb{\tilde{\X}\vert\X}{\tilde{\x}\vert\x} \prob{\X\vert\Y}(\x \vert \y) \;\mathrm{d}\tilde{\x}\,\mathrm{d}\x \right] \\
			&\hphantom{= } +  \left[ \int \int \Big< \nabla_{\tilde{\x}} \log  \prob{\tilde{\X}\vert\X}\!(\tilde{\x}\vert\x), \nabla_{\tilde{\x}} \log  \prob{\tilde{\X}\vert\X}\!(\tilde{\x}\vert\x) \Big> \; \probb{\tilde{\X}\vert\X}{\tilde{\x}\vert\x} \prob{\X\vert\Y}(\x \vert \y) \;\mathrm{d}\tilde{\x}\,\mathrm{d}\x \right] \Bigg\}\\
			&\hphantom{= }+ \mathcal{C}_1 - \mathcal{C}_2 \\
			&= \frac{1}{2} \Bigg\{ \int \int \bigg[ \Big< \sm(\tilde{\x}, \y; \thetaa), \sm(\tilde{\x}, \y; \thetaa)\Big> - 2 \Big< \sm(\tilde{\x}, \y; \thetaa), \nabla_{\tilde{\x}}  \log \probb{\tilde{\X}\vert\X}{\tilde{\x}\vert\x} \Big> \\
			&\hphantom{= \Bigg\{ }+  \Big< \nabla_{\tilde{\x}} \log  \prob{\tilde{\X}\vert\X}\!(\tilde{\x}\vert\x), \nabla_{\tilde{\x}} \log  \prob{\tilde{\X}\vert\X}\!(\tilde{\x}\vert\x) \Big> \bigg] \probb{\tilde{\X}\vert\X}{\tilde{\x}\vert\x} \prob{\X\vert\Y}(\x \vert \y) \;\mathrm{d}\tilde{\x}\,\mathrm{d}\x \Bigg\} + \mathcal{C}_1 - \mathcal{C}_2 \\
			&= \frac{1}{2} \int \int \big\lVert \sm(\tilde{\x}, \y; \thetaa) - \nabla_{\tilde{\x}} \log  \prob{\tilde{\X}\vert\X}\!(\tilde{\x}\vert\x) \big\rVert^2_2 \; \probb{\tilde{\X}\vert\X}{\tilde{\x}\vert\x} \prob{\X\vert\Y}(\x \vert \y) \;\mathrm{d}\tilde{\x}\,\mathrm{d}\x + \underbrace{\mathcal{C}_1 - \mathcal{C}_2}_{\mathcal{C}} \\
			&= \frac{1}{2} \Exp_{\X \sim \prob{\X\vert\Y}} \Bigg[ \Exp_{\tilde{\X} \sim \prob{\tilde{\X}\vert\X}} \Big[ \lVert  \sm(\tilde{\x}, \y; \thetaa) - \nabla_{\tilde{\x}} \log  \probb{\tilde{\X}\vert\X}{\tilde{\x}\vert\x}\rVert_2^2 \Big] \Bigg] + \mathcal{C}. 
		\end{split}
	\end{equation}
	This completes the derivation of \Cref{eq:DSM-2} from \Cref{eq:dsm1} using \Cref{eq:convolution}.
	
		\section{Details of Monte Carlo simulation for estimating posterior statistics for the inverse problem in \Cref{subsubsec:elasto}}\label{app:MCS}
		
		Corresponding to a measurement $\hat{\y}$, we estimate the posterior mean as follows:
		\begin{equation}
			\bar{\bm{\mu}} = \Exp_{\X \sim \prob{\X\vert\Y}} \left[ \x \right] = \frac{\sum_{i}^{\Ns} w_i \x^{(i)}}{\sum_{i}^{\Ns} w_i } ,
		\end{equation}
		where $\x^{(i)}$ are independent realizations of $\X$ sampled from the latent prior distribution, $\Ns$ is the sample size, and $w_i$ are the importance weights. Similarly, we estimate the posterior's covariance matrix as follows:
		\begin{equation}
			\mathrm{Var}_{\X \sim \prob{\X\vert\Y}} \left[ \x \right] = \frac{\sum_{i}^{\Ns} w_i \left( \x^{(i)} - \bar{\bm{\mu}} \right)\transpose \left( \x^{(i)} - \bar{\bm{\mu}} \right)}{\sum_{i}^{\Ns} w_i }. 
		\end{equation}
		The diagonal entries of the posterior's covariance matrix provides the pixel-wise variance. Since the measurement noise is additive and Gaussian with the standard deviation known \latin{a priori}, we use the likelihood function to compute $w^i$ as follows:
		\begin{equation}
			w_i =  (2\pi )^{-\Ny/2}\det(\Sigm_{\eta})^{-1/2}\,\exp \left(-{\frac {1}{2}}(\Fii(\x^{(i)}) - \hat{\y})^{\mathrm {T} }{\Sigm_{\eta}}^{-1}(\Fii(\x^{(i)}) - \hat{\y})\right),
		\end{equation}
		where $\Sigm_{\eta} = \sigma^2_{\eta} \mathbb{I}_{\Ny}$ is covariance matrix for the Gaussian likelihood function, and $\Fii$ is the computational model that predicts the full-field response from the spatial distribution of the shear modulus is $\x^{(i)}$.

		\section{Additional details for the optical coherence elastography application in \Cref{subsubsec:TS}}\label{app:TS_details}
		
		\subsection{Background}\label{appsubsec:TS_background}
		OCT is a technique that leverages optics, analogous to acoustics in ultrasound imaging, to produce high-resolution tomograms of tissue. A functional extension of OCT that enables imaging based on mechanical contrast is OCE~\cite{kennedy2014review}, which involves capturing the phase data alongside intensity measurements of the interference of light backscattered from the specimen with light reflected from a reference mirror at varied axial depths. One variant of OCE is phase-sensitive compression OCE~\cite{kennedy2014optical}, wherein the phase shifts observed between B-scans (cross-sectional images) of the specimen as it undergoes compressive loading quantify the specimen's deformation along the optical axis. The mechanical properties of the specimen can be inferred from these phase shifts because they encode within them information about the displacement component along the optical axis. We use a cSDM to solve the inverse problem of determining elasticity profiles, specifically the spatial distribution of the Young's modulus $\X$ of tumor spheroids from noisy phase difference measurements $\Y$. 
		
		\subsection{Experimental setup}\label{appsubsec:TS_exp_details}
		We provide a brief summary of the experimental setup here and the reader is referred to  \cite{foo2024tumor} for more details. In the experiment, a custom-built spectral-domain OCM system in dual-arm configuration images tumor spheroids embedded in hydrogel specimens from the top. Following the objective lens, the sample arm contains, from top to bottom: a rigid glass coverslip acting as an imaging window, a 300~$\mu$m thick sample containing tumor spheroids, a 1~mm thick compliant silicone layer, and a piezoelectric actuator to apply compressive displacement. The purpose of the compliant layer is to provide surface stress measurements for mechano-microscopy~\cite{mowla2022subcellular}; we do not utilize this data in this study. Two volumetric images at different compression levels were acquired at each location to produce a volumetric phase difference image. The plane corresponding to the central cross-section of the tumor spheroid was identified in the volumetric image and was cropped to a 256$\times$256~$\mu$m\textsuperscript{2} window around the spheroid. This two-dimensional phase difference image was used as measurement in solving the inverse problem.
		
		\subsection{Specimen preparation}\label{appsubsec:TS_specimen}
		The tumor spheroid specimens were prepared by encapsulating MCF7 (non-metastatic breast cancer) cells within a hydrogel matrix composed of lyophilized gelatin methacryloyl (GelMA). The cells were cultured until they formed spheroids. Each tumor spheroid contains multiple cells, and each cell domain can be broadly classified into two components: the cytoplasm and the nucleus (see \Cref{fig:ts_schematic}). We take inspiration from this morphology to develop the parametric prior for this study.
		
		\subsection{Parametric prior}\label{appsubsec:TS_prior}
		We model the spheroid as an ellipse, and straight lines subdivide the spheroid into cells. Each cell contains a circle representing a nucleus, surrounded by the cytoplasm of the cell. We treat the following parameters as truncated normal random variables: spheroid width, spheroid aspect ratio, spheroid position, number of nuclei per unit area, nuclei radii, nuclei elasticity, cytoplasm elasticity, and hydrogel elasticity. \Cref{tab:ts_prior} provides details on these parameter's distributions, which we derive from preliminary analyses of spheroid OCM images. We determine the nuclei locations (defined by their center coordinates) using a custom algorithm that iteratively accumulates nuclei coordinates by creating a list of valid coordinates for the current nucleus to be positioned, randomly selecting a coordinate from this list for the current nucleus, removing the coordinates occupied by the current nucleus from the list of valid coordinates for the next nucleus, and repeating until the required number of nuclei for the spheroid is reached or there are no more valid coordinates for the next nucleus. We constrain the sizes and positions of the nuclei to ensure a minimum spacing of 2 $\mu$m between nuclei and between nuclei and the ellipse boundary. We subsequently determine the cell boundaries using a power diagram, also known as a Laguerre–Voronoi diagram~\cite{imai1985voronoi}.
		\begin{table}[H]
			\centering
			\caption{Random variables comprising the parametric prior for the tumor spheroid application}
			\label{tab:ts_prior}
			\setlength\tabcolsep{0pt}
			\begin{tabular*}{\textwidth}{@{\extracolsep{\fill}} lccc}
				\toprule
				Parameter & Variable & Minimum & Maximum \\
				\toprule
				Spheroid width ($\mu$m) & $\mathcal{N}(105,8.33)$ & 80 & 130 \\
				Spheroid height to width ratio & $\mathcal{N}(0.90,0.0167)$ & 0.85 & 0.95\\
				Spheroid (center) horizontal position ($\mu$m) & $\mathcal{N}(0,13.33)$ & -40 & 40 \\
				Spheroid (center) vertical position ($\mu$m) & $\mathcal{N}(0,13.33)$ & -40 & 40 \\
				Number of nuclei per square area ($\text{mm}^{-2}$) & $\mathcal{N}(4750,250)$ & 4000 & 5500 \\
				Radius of nuclei ($\mu$m) & $\mathcal{N}(7.5,0.5)$ & 6 & 9\\
				Young's modulus of nuclei (kPa) & $\mathcal{N}(32.5,5.833)$ & 15 & 50 \\
				Young's modulus of cytoplasm (kPa) & $\mathcal{N}(10,1.67)$ & 5 & 15\\
				Young's modulus of hydrogel (kPa) & $\mathcal{N}(7.5,0.833)$  & 5 & 10\\
				Displacement applied to top edge ($\mu$m) & $\mathcal{N}(-2.2,0.6)$  & $-0.4$ & $-4$ \\ 
				\bottomrule
			\end{tabular*}
			\begin{minipage}{\textwidth}
				\small{*In this example, the normal distributions were truncated at three standard deviations from the mean, with truncation limits explicitly defined by the specified minimum and maximum values.}
			\end{minipage}
		\end{table}
		
		\subsection{Forward model}\label{appsubsec:TS_fwdmodel}
		Given a realization $\x$ of the Young's modulus field, we simulate uniaxial compression on the specimen via FEA using ABAQUS~\cite{simulia2022user} to obtain the vertical displacement field (third column in \Cref{fig:ts_schematic}). To mitigate edge effects, we initially expand each sample to an 800$\times$800~$\mu\text{m}^2$ domain by padding with hydrogel on all four sides. We displace the top edge downward while the bottom-left corner remains fixed, and the bottom edge is constrained vertically. We sample the displacement on the top edge from a truncated normal distribution; see \Cref{tab:ts_prior}. The magnitude of the displacement of the top edge ensures that the corresponding strain values are within the order of a few millistrains, consistent with observations from mechano-microscopy experiments. For this analysis, given the low levels of strain involved, we assume that the specimen is linear and elastic (\Cref{eq:linear_constitutive_law}), under plane strain conditions (\Cref{eq:equilibrium_equation}), and nearly incompressible with Poisson's ratio uniformly set to 0.49. We develop a CAD model for the specimen (second column in \Cref{fig:ts_schematic}), which we subsequently use to develop the FEA mesh geometry consisting of quadrilateral elements for the main region and triangular elements for the padded regions. Once the FEA is complete, we obtain 2D arrays of the Young's modulus $E$ and vertical displacement $u_2$ by interpolating the main 256$\times$256~$\mu\text{m}^2$ domain window to a 256$\times$256 Cartesian grid.  We convert the displacement fields to phase difference fields $\Delta \phi$ using the following equation:
		\begin{equation}
			\Delta \phi =  \frac{4\pi n_r u_2}{\lambda_0} \;,
		\end{equation}
		where $n_r=1.4$ is the refractive index of the hydrogel, and $\lambda_0 = 800$~nm is the central wavelength of the OCM light source. 
		
		Subsequently, we add non-Gaussian measurement noise $\phi_\eta$ to the phase difference field to obtain a noisy measurement image (fourth column in \Cref{fig:ts_schematic}). The noise in the phase difference varies with depth and is modeled to follow a probability law characterized by the following density function~\cite{foo2024tumor}:
		\begin{equation}\label{appeq:TS_noisemodel}
			\text{p}_{\Phi_\eta}(\phi_\eta) = \frac{1}{2\pi}\exp{\left(-\frac{k^2}{2}\right)} + \sqrt{\frac{1}{8\pi}}k \exp{\left(-\frac{k^2}{2}\sin^2{\phi_\eta}\right)}\cos{\phi_\eta} \left[1 +\operatorname{erf}\left(\frac{k\cos{\phi_\eta}}{\sqrt{2}}\right)\right],
		\end{equation}
		where $k$ is related to the variance of this distribution. The variance varies with depth along the optical axis in a B-scan due to optical attenuation and the confocal function of the objective lens, and was determined as a function of the depth by curve fitting based on experimental calibration conducted with hydrogel specimens~\cite{foo2024tumor}. After the addition of noise, the phase difference is wrapped to ensure it remains within the range $(-\pi,\pi]$. We obtain the final wrapped phase difference image $\Delta \phi_{\rm w}$ using the following formula
		\begin{equation}
			\Delta \phi_{\rm w} =\arg \left(\exp\left[i\left(\Delta \phi + \phi_\eta\right)\right]\right).
		\end{equation}
		This wrapped and noisy phase difference field is the realization $\y$ of the measurements corresponding to the realization $\x$ of the Young's modulus field.
	
\end{document}